\pdfoutput=1

\documentclass[11pt]{article}

\usepackage[]{EMNLP2022}

\usepackage{times}
\usepackage{latexsym}
\usepackage{hyperref}
\usepackage{url}
\usepackage{caption}
\usepackage{graphicx}
\usepackage{tabularx}
\usepackage{booktabs}
\usepackage{multicol, multirow}
\usepackage{wrapfig}
\usepackage{subfigure}
\usepackage{enumitem}
\usepackage{color,soul}
\usepackage{longtable}

\setul{0.6ex}{0.2ex}

\usepackage{microtype}

\newcommand{\DatasetName}{\textsc{BookSum}}

\title{\DatasetName: A Collection of Datasets for Long-form Narrative Summarization}

\newcommand*{\affaddr}[1]{#1}
\newcommand*{\affmark}[1][*]{\textsuperscript{#1}}

\author{
  \textbf{Wojciech Kry\'sci\'nski}\affmark[$\dagger$]
  \quad \textbf{Nazneen Rajani}\affmark[$\mathsection$]
  \quad \textbf{Divyansh Agarwal}\affmark[$\dagger$] \\
  \quad \textbf{Caiming Xiong}\affmark[$\dagger$]
  \quad \textbf{Dragomir Radev}\affmark[$\dagger\ddagger$] \\
  \affaddr{
  \affmark[$\dagger$]Salesforce Research
  \quad
  \affmark[$\mathsection$]Huggingface
  \quad 
  \affmark[$\ddagger$]Yale University} \\
  \{kryscinski, divyansh.agarwal, cxiong\}@salesforce.com \\ nazneen@huggingface.co \\ dragomir.radev@yale.edu
}

\date{}

\begin{document}
\maketitle
\begin{abstract}
The majority of existing text summarization datasets include short-form source documents that lack long-range causal and temporal dependencies, and often contain strong layout and stylistic biases.
While relevant, such datasets will offer limited challenges for future text summarization systems.
%
We address these issues by introducing \DatasetName, a collection of datasets for long-form narrative summarization.
Our dataset covers documents from the literature domain, such as novels, plays and stories, and includes highly abstractive, human written summaries on three levels of granularity of increasing difficulty: paragraph-, chapter-, and book-level.
The domain and structure of our dataset poses a unique set of challenges for summarization systems, which include: processing very long documents, non-trivial causal and temporal dependencies, and rich discourse structures.
To facilitate future work, we trained and evaluated multiple extractive and abstractive summarization models as baselines for our dataset.
\end{abstract}

\section{Introduction}\label{sec:introduction}
Text summarization aims at condensing long documents into a short, human-readable form which contains only the salient parts of the source.
Leveraging the cutting-edge findings in natural language processing, such as multi-task learning methods~\citep{Raffel:19}, pre-training strategies~\citep{Zhang:19b}, and memory-efficient architectures~\citep{Zaheer:20}, text summarization has seen substantial progress.
The majority of papers published in the field focus on summarizing newswire documents from popular datasets, such as CNN/DailyMail~\citep{Nallapati:16}. 
Other domains gaining interest of the research community are scientific and legal documents, with notable datasets being Arxiv/PubMed~\citep{Cohan:18} and BigPatent~\citep{Sharma:19}.
While the performance of state-of-the-art methods on those datasets is impressive, the mentioned domains have inherent shortcomings and thus pose limited challenges for future generations of text summarization systems.
First, the length of summarized documents is limited, ranging from a few hundred words in case of news articles, to a few pages for scientific documents and patent applications. 
%
In most cases, such short-form documents can be quickly read by humans, thus limiting the practical value of automatic summarization systems.
%
%
Second, the domains under consideration impose strict requirements regarding the document's layout and stylistic features\footnote{\url{owl.purdue.edu/owl/purdue_owl.html}}.
Statements follow a logical order and all facts are offered explicitly, leaving limited space for interpretation and reasoning.
Additionally, such constraints, can introduce layout biases into the datasets which later dominate the training signal of the summarization systems.
The lead-bias present in news articles being one example of such effects~\citep{Kedzie:18, Kryscinski:19}.
Third, documents in the mentioned domains lack long-range causal and temporal dependencies, and rich discourse structures.
Due to the limited length and fact-centric style of writing, most causal dependencies span only a few paragraphs, temporal dependencies are organized in a monotonic fashion where newly introduced facts refer only to previously stated information, and document lacks features such as parallel plot lines.

\begin{figure}[tb]
    \centering
    \includegraphics[width=0.95\linewidth]{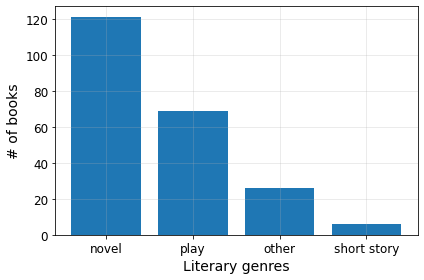}
    \caption{
    Distribution of literary genres included in \DatasetName.
    The \textit{other} category includes works such as autobiographies, poems, and political books.
    }
    \label{fig:book-categories}
\end{figure}

In this work we address the shortcomings of existing datasets and introduce~\DatasetName, a collection of data resources for long-form narrative summarization~\citet{Ladhak:20}.
The data covers documents from the literature domain, including stories, plays, and novels (Fig.~\ref{fig:book-categories}), each provided with a highly abstractive, human-written summary.
Leveraging the characteristics of fiction writing, \DatasetName~introduces a set of new challenges for summarization systems: processing long-form texts ranging up to hundreds of pages, understanding non-trivial causal and temporal dependencies spread out through the entirety of the source, handling documents with rich discourse structure which include parallel plots or changes between narration and dialogue, and generating highly abstractive and compressive summaries.
Solving such challenges will require progress in both automatic document understanding and processing of long inputs.
To support incremental progress, the \DatasetName~collection includes examples on three levels of granularity with increasing difficulty:
1) paragraph-level, with inputs consisting of hundreds of words and short, single-sentence summaries,
2) chapter-level, with inputs covering several pages and multi-sentence summaries,
3) book-level, with inputs spanning up to hundreds of pages and multi-paragraph summaries.
The hierarchical structure of the dataset, with aligned paragraph, chapter, and book-level data, makes it a viable target for single- and multi-document summarization approaches.

To demonstrate the new set of challenges for text summarization models introduced by~\DatasetName~and lay the groundwork for future research, we evaluated several state-of-the-art extractive and abstractive summarization architectures on the newly introduced task.
We share the data preparation scripts here:~\url{https://github.com/salesforce/booksum}.

\section{Related Work}\label{sec:related-work}
The availability of digital documentation has translated into a number of novel, large-scale datasets for text summarization that span a variety of domains.
In the news domain, \citet{Sandhaus:08} introduced a corpus of news articles from the New York Times magazine with summaries written by library scientists.
\citet{Nallapati:16} collected articles from the CNN and DailyMail portals with multi-sentence article highlights repurposed as reference summaries.
\citet{Narayan:18} aggregated articles from the BBC website with highly abstractive, single sentence summaries.
\citet{Grusky:18} introduced a dataset spanning 38 news portals, with summaries extracted from the websites metadata.
In the academic article domain, \citet{Cohan:18} collected scientific articles from the Arxiv and PubMeb article repositories and used paper abstracts as reference summaries.
\citet{Wang:20} aggregated a set of articles in the medical domain related to the Covid-19 pandemic, also using paper abstracts as reference summaries.
\citet{Hayashi:20} introduced a multi-domain collection of scientific articles each with two associated summaries, one covering the article's contributions, the other explaining the context of the work.
Related to dialogue summarization, \citet{Pan:18} repurposed image captioning and visual dialogue datasets to create a summarization dataset containing conversations describing an image, with image captions considered the summaries.
\citet{Gliwa:19} introduced a corpus of conversations between hired annotators designed to mimic interactions on a messaging application with human written summaries.
In the legal domain, \citet{Sharma:19} has collected a. large collection of patent filings with associated, author-written invention descriptions.
%

Despite the increased interest in the broader field of text summarization, little work has been done in summarizing stories and novels.
In \citet{Kazantseva:06}, the authors focused on generating extractive overviews of short works of fiction. 
The work proposed two modeling approaches, one utilizing decision trees the other based on a manually designed system of rules with experiments conducted on a set of 23 short stories.
%
%
%
\citet{Mihalcea:07}~introduced the task of book summarization along with a set of resources and baselines.
The authors collected and curated a set of 50 books from the Gutenberg Project with two human-written summaries associated with each book collected from online study guides.
%
%
%
%
More recently, \citet{Zhang:19} tackled the problem of generating character descriptions based on short fiction stories.
The authors collected a dataset of stories with associated, author-written summaries from online story-sharing platforms and proposed two baseline methods for solving the task. 
%
\citet{Ladhak:20} explored the problem of content selection in novel chapter summarization.
The authors studied different approaches to aligning paragraphs from book chapters with sentences from associated summaries and created a silver-standard dataset for extractive summarization. 
The work also studied the performance of extractive models on the task.
%
%

Our work extends the efforts made by \citet{Ladhak:20}.
The \DatasetName~corpus prioritizes abstractive summarization and offers aligned data on three levels of granularity (paragraph, chapter, full-book), substantially increasing the number of available examples.
We also benchmark the performance of state-of-the-art extractive and abstractive methods on all introduced data subsets.
\begin{table*}[th]
    \centering
    \small
    \begin{tabular}{lrrrrrr} 
    \toprule
    \textbf{Dataset} & \textbf{\# Docs.} & \textbf{Coverage} & \textbf{Density} &  \textbf{Comp. Ratio} & \multicolumn{2}{c}{\textbf{\# Tokens}} \\
    & & & & & Source & Summary \\
    \midrule
    Arxiv/PubMed & 346,187 & 0.87 & 3.94 & 31.17 & 5179.22 & 257.44 \\
    BigPatent & 1,341,306 & 0.86 & 2.38 & 36.84 & 3629.04 & 116.66 \\
    CNN/DM & 311,971 & 0.85 & 3.47 & 14.89 & 803.67 & 59.72 \\
    Newsroom & 1,212,739 & 0.83 & 9.51 & 43.64 & 799.32 & 31.18 \\
    XSum & 226,677 & 0.66 & 1.09 & 19.25 & 438.43 & 23.89 \\
    NovelChapters* & 8,088 & - & - & - & 5,165 & 372 \\ \midrule
    \DatasetName~Paragraph (ours) & 146,532 & 0.50 & 0.92 & 6.47 & 159.55 & 40.59 \\
    \DatasetName~Chapter (ours) & 12,630 & 0.78 & 1.69 & 15.97 & 5101.88 & 505.42 \\
    \DatasetName~Full (ours) & 405 & 0.89 & 1.83 & 126.22 & 112885.15 & 1167.20 \\
    \bottomrule
    \end{tabular}
    
    \caption{
    Statistics of the \DatasetName~data collection compared with other popular text summarization datasets.
    *NovelChapters dataset~\citep{Ladhak:20} could not be reliably reproduced at the time of writing of this work, the numbers were copied from the original paper.
    }
    \label{tab:data-stats}
\end{table*}

\section{Dataset}\label{sec:dataset}

In this section we describe the data sources and pre-processing steps taken to create the \DatasetName~data collection and conduct an in-depth analysis of the collected resources.

\subsection{Data Collection}\label{ssec:data-collection}

\paragraph{Data Sources} Despite the popularity of books in electronic format, aggregating and sharing literature pieces is a non-trivial task due to the copyright law protecting such documents.
The source documents available in \DatasetName~were collected from the Project Gutenberg public-domain book repository\footnote{US edition: \url{https://www.gutenberg.org/}} and include plays, short stories, and novels of which copyrights have expired.
Associated summaries were collected using content provided by the Web Archive\footnote{\url{https://web.archive.org/}}.
The summary data includes both book- and chapter-level summaries. 

\paragraph{Data Acquisition}
Source texts were downloaded in plain text format in accordance with Project Gutenberg's guidelines\footnote{\url{https://www.gutenberg.org/policy/robot_access.html}}.
The data collection contains texts exclusively from the US edition of Project Gutenberg.
Summaries were collected using content provided by the Web Archive and processed using the BeautifulSoup library\footnote{\url{https://crummy.com/software/BeautifulSoup/}}.
Collecting summaries from several independent sources with small content overlaps between them resulted in certain texts having multiple associated summaries.
Upon manual inspection, substantial stylistic differences were found between the related summaries, thus such coverage overlap was considered advantageous for the dataset.

\paragraph{Data Cleaning \& Splitting}
To ensure high quality of the data, both the source texts and summaries were cleaned after collection.
Metadata containing author, title, and publisher information was removed from source files.
The documents were manually split into individual chapters to accommodate chapter-level summarization.
Due to the unstructured nature of plain text files, heuristic approaches were used to extract chapter content.
Initial, automatic chapterization was done using the regex-based Chapterize tool\footnote{\url{https://github.com/JonathanReeve/chapterize}}.
However, an inspection of outputs revealed many partially processed and unprocessed files, such instances were chapterized manually by the authors of this work.
Paragraph-level data was obtained by further splitting the extracted chapter into individual paragraphs based on a white-character pattern.
Short paragraphs and dialogue utterances were aggregated to form longer paragraphs. 
Collected summaries were also inspected for scraping artifacts and superfluous information.
Regular expressions were used to remove leftover HTML tags, author's notes, and analysis parts that were not directly related to the content of the summary.

\paragraph{Data Pairing} Source texts and associated summaries were collected independently of each other and required alignment.
The pairing procedure was conducted in phases, starting with coarse-grained full-text alignments and ending with fine-grained paragraph alignments, with each phase involving automatic alignments followed by manual inspection and fixes.
Full texts were paired with summaries based on title matches and later verified by matching author names.
To accommodate automatic alignment, titles were normalized into a common format with lower-case letters and all punctuation characters removed.
Chapter alignments were based on chapter metadata, extracted during source text chapterization, and chapter titles collected from online study guides.
Similar to full-text titles, chapter names were transformed to a common format with chapter names lower-case and cleaned from punctuation characters, and chapter numbers translated to roman numerals.
Paragraph-level alignments were computed between paragraphs extracted from chapters and individual sentences of chapter-level summaries.
Following a two step process introduced by~\citet{Ladhak:20}, the alignment process was preceded by a human-based study aimed at finding an optimal alignment strategy, with its details presented in Appendix~\ref{app:data-align}.
With the insights from the study, paragraph-sentence similarities were computed using a SentenceTransformer~\citep{Reimers:19}, and leveraged a stable matching algorithm~\citep{Shapley:62} to obtain the final alignments.
All examples on the chapter- and book-level, and a random subset of examples on the paragraph-level were manually inspected to ensure high quality of data.
Quantitative verification of alignment quality is discussed in Appendix~\ref{app:align-quality}.

\paragraph{Data Splits}
The data was split into training, validation, and test subsets in a 80/10/10\% proportion.
To prevent data leakage between data subsets, the splits were assigned per book title, meaning that all paragraph, chapter, and full-book examples belonging to the same book title were assigned to the same data split.
For consistency with the dataset introduced by~\citet{Ladhak:20}, all titles overlapping between the two datasets were assigned to the same splits. 
Remaining titles were assigned to splits at random following the predefined size proportions.
The data collection and pre-processing pipeline is visualized in Figure~\ref{fig:pipeline-figure} in the Appendix~\ref{app:data-pipeline}.

\subsection{Data Analysis}
\begin{figure*}[tb]
\centering
    \subfigure[Salient unigram distribution]{{\includegraphics[width=0.49\linewidth]{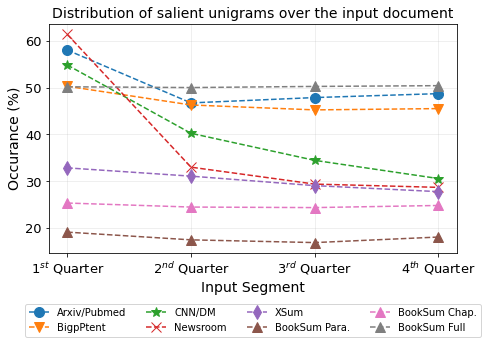}}}%
    \subfigure[Percentage of novel $n$-grams]{\includegraphics[width=0.49\linewidth]{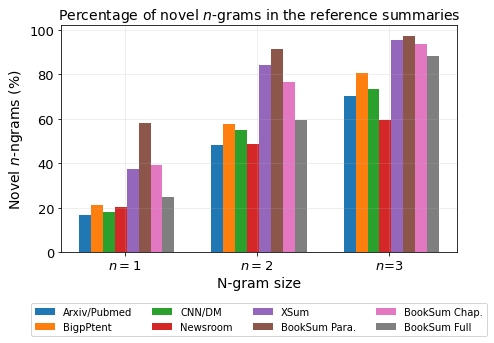}}%
    \caption{The datasets statistics of \DatasetName~and previously introduced datasets.
    Figure (a) shows the salient unigram distribution over 4 equally sized segments of the source documents.
    Figure (b) shows the percentage of novel $n$-grams in the reference summaries when compared with the source documents.
    }
    \label{fig:data-stats-figs}
\end{figure*}

\paragraph{Data Statistics}
The data collection and matching process described in Section~\ref{ssec:data-collection} yielded 217 unique book titles with a total of 6,327 book chapters.
After the pre-processing and alignment steps, the \DatasetName~collection contains 146,532 paragraph-level, 12,630 chapter-level, and 405 book-level examples.
Figure~\ref{fig:book-categories} shows the distribution of literary genres in our corpus.
Following~\citet{Grusky:18}, we computed statistics of the \DatasetName~collection and compared them with other popular summarization datasets in Table~\ref{tab:data-stats}.
Coverage and density, which measure the extractive span similarity between source and summary, indicate that while the extractiveness of summaries increases from 0.5 and 0.92 for paragraphs to 0.89 and 1.83 for full-books, the summaries are still highly abstractive when compared to other datasets, such as CNN/DM or Newsroom.
Relatively low coverage and density scores for paragraph-level alignments might partially be an artifact of the heuristic approach to aligning the data.
The lengths of source and summary texts substantially increases across data granularity.
Paragraph-level data includes short documents with an average of 159 words which fit within the limitations of existing models, chapter-level examples contain texts with average of over 5000 words, which are longer than in most of existing datasets and go beyond limitations of many state-of-the-art methods~\citep{Liu:19b}, while book-level examples contain inputs with over 110,000 words on average, which are orders of magnitude longer than any document previously used in NLP tasks.
While long source documents create computational challenges for encoding components of models, the associated summaries on chapter- and book-level are also much longer than in any other dataset, thus creating challenges for the generative component of summarization methods.

\paragraph{Salient Content Distribution}
To assess the difficulty of content selection in our datasets we measure the distribution of salient unigrams in the source texts~\citep{Sharma:19}.
The distribution is computed as the percentage of salient unigrams in four equally sized segments of the source text, where salient unigrams are words appearing in the associated summaries after removing stopwords.
As shown in Figure~\ref{fig:data-stats-figs}~(a), all subsets of the \DatasetName~dataset have a relatively even distribution of salient words across all four segments of the source documents.
This suggests that to generate high quality paragraph, chapter, or book summaries models will have to use the entire source document instead of only relying on parts of it.
In comparison, other datasets, such as CNN/DM, Newsroom, or Arxiv/Pubmed, contain strong layout biases where the majority of salient words appear in the first quarter of the source documents.

\paragraph{Summary Abstractiveness}
To quantify the abstractiveness of summaries in \DatasetName~we measured the percentage of $n$-grams from summaries not appearing in the associated source document~\citep{See:17}.
Results presented in Figure~\ref{fig:data-stats-figs}~(b) show that \DatasetName~contains highly abstractive summaries across all measured $n$-gram sizes.
The highest ratio of novel $n$-grams in \DatasetName~was found for the paragraph-level alignments, followed by chapter-level data and full-books.
Results also indicate that our dataset is substantially more abstractive than most previous datasets, with the exception of XSum.
High scores for trigrams also indicate that summaries included in \DatasetName~do not contain long extractive spans, which aligns with the Density statistics shown in Table~\ref{tab:data-stats}.

\paragraph{Qualitative Study} 

For a deeper understanding of the data beyond quantitative evaluation, we manually analyzed subsets of \DatasetName.
First we compared summaries on different levels of granularity assigned to the same title. 
Summaries on the chapter- and book-level partially overlap in the summarized content, however substantially differ in the level of detail with which they cover the content.
This relation could be leveraged for training models in a hierarchical fashion, from shorter to longer source texts~\citep{Li:15}.
Next, we compared summaries coming from different sources which were aligned with the same book or chapter.
We noticed that the summaries had high semantic and low lexical overlap, meaning that they covered the same content of the summarized documents, but were written in a unique way.
Such examples contain useful training signal for abstractive summarization models.
Table~\ref{tab:source-examples} shows examples of chapter summaries of \textit{"Sense and Sensibility"}.

\section{Experiments}\label{sec:experiments}
\begin{table*}[tb]
    \centering
    \resizebox{\linewidth}{!}{%
    \small
    \begin{tabular}{lrrrrrrrrrrrrrrr} 
    \toprule
    & \multicolumn{5}{c}{\DatasetName-Paragraph} & \multicolumn{5}{c}{\DatasetName-Chapter} & \multicolumn{5}{c}{\DatasetName-Book} \\
    \cmidrule(lr){2-6} 
    \cmidrule(lr){7-11}
    \cmidrule(lr){12-16}
    \textbf{Models} & $\textrm{R-1}_{f_1}$ & $\textrm{R-2}_{f_1}$ & $\textrm{R-L}_{f_1}$ & $\textrm{BS}_{f_1}$ & $\textrm{SQA}_{f_1}$ & $\textrm{R-1}_{f_1}$ & $\textrm{R-2}_{f_1}$ & $\textrm{R-L}_{f_1}$ & $\textrm{BS}_{f_1}$ & $\textrm{SQA}_{f_1}$ & $\textrm{R-1}_{f_1}$ & $\textrm{R-2}_{f_1}$ & $\textrm{R-L}_{f_1}$ & $\textrm{BS}_{f_1}$ & $\textrm{SQA}_{f_1}$ \\

    \midrule
    & & \multicolumn{11}{c}{Heuristcs} \\
    \midrule
    Lead-3 & 17.99 & 3.25 & 12.80 & 0.085 & 22.18 & 14.32 & 2.23 & 8.59 & -0.008  &  19.28 & 6.50 & 0.89 & 4.34 & -0.060 & 24.18 \\
    Random Sentences & 17.56 & 3.03 & 12.32 & 0.080 & 19.99 & 12.54 & 1.32 & 7.43  & -0.029 & 7.86 & 5.26 & 0.48 & 3.38 & -0.119 & 6.28 \\
    Extractive Oracle & 27.90 & 7.22 & 20.96 & 0.172  & 17.33 & 42.36 & 9.83 & 20.91 & 0.153 & 17.16 &  46.30 & 9.25  & 17.77  & 0.076 & 17.90 \\ 
    \midrule
    
    & & \multicolumn{11}{c}{Extractive Models} \\
    \midrule
    CNN-LSTM  & 16.71  & 2.93 & 12.84 & 0.096 & 7.17 & 32.50 & 5.51  & 13.91  & 0.081 & 3.81 &  38.79 & 6.90 & 14.28 & 0.033 & 3.09   \\
    BertExt & 14.55 & 2.29 & 10.52 & 0.053 & 8.12 & 32.06 & 5.37 & 13.68 & 0.076 & 8.61 & 36.32 & 6.04 & 13.29 & 0.020 & 9.64 \\
    MatchSum & 19.13 & 3.38 & 14.17 & 0.125 & 16.47 & 30.97 & 5.34  & 13.23  & 0.082 & 16.83 & 33.76  & 5.58  & 12.68  & 0.012 & 24.11 \\
    
    \midrule
    & & \multicolumn{11}{c}{Abstractive Models} \\
    \midrule
    BART~\textsubscript{zero-shot} & 17.66 & 2.29 & 13.15 & 0.132 & 12.03 & 31.54 & 5.25 & 14.20 & 0.088 & 11.80 & 35.35 & 5.36 & 12.93 &  0.021 & 11.90 \\
    T5~\textsubscript{zero-shot} & 19.84 & 3.61 & 14.05 & 0.096 & 19.07 & 31.28 & 5.25 & 13.05 & 0.075 & 19.00 & 35.50 & 5.60 & 12.04 & 0.011 & 20.43 \\
    PEGASUS-large~\textsubscript{zero-shot} & 16.35 & 2.61 & 12.01 & 0.093 & 14.34 & 30.61 & 4.51 & 13.07 & 0.077 & 14.47 & 34.16 & 5.05 & 12.40 & 0.021 & 20.29 \\
    
    \midrule
    BART~\textsubscript{fine-tuned} & 22.79 & 5.09 & 17.70 & 0.211 & 13.96 & 37.51 & 8.49 & 17.05 & 0.156 & 13.64 & 38.71 & 7.59 & 13.65 & 0.051 & 14.89 \\
    T5~\textsubscript{fine-tuned} & 21.07 & 4.87 & 16.96 & 0.212 & 13.93 & 36.60 & 7.93 & 16.79 & 0.152 & 13.65 & 39.87 & 8.01 & 13.99 &  0.074 & 15.13\\
    PEGASUS-large~\textsubscript{fine-tuned} & 20.32 & 4.56 & 16.22 & 0.194 & 10.23 & 35.86 & 7.47 & 16.10 & 0.132 & 11.54 & 36.03 & 7.23 & 12.88  & 0.050 & 14.85 \\

    \bottomrule
    \end{tabular}
    }%
    \caption{
    Performance of baseline models on the Paragraph, Chapter, and Full-Book subsets of \DatasetName~evaluated with automatic metrics: ROUGE-n (R-n), BERTScore (BS), and SummaQA (SQA).
    }
    \label{tab:rouge-scores}
\end{table*}

To motivate the challenges posed by the \DatasetName~corpus, we study the performance of multiple baseline models, both extractive and abstractive, on the different levels of alignment: paragraph, chapter and books.
We refer to these levels of alignment as \DatasetName-Paragraph, \DatasetName-Chapter, and \DatasetName-Book accordingly.
%
%

\subsection{Baseline Models}\label{ssec:baseline-models}
\paragraph{Lead-3}\citep{See:17}
is an extractive heuristic where the first three sentences from the source document are treated as the summary.
Despite its simplicity, Lead-3 is a strong baseline for domains which show layout biases, such as newswire.

\paragraph{Random Sentences}
follows the Lead-3 heuristic and extracts 3 sentences sampled at random from the source document.
It represents the performance of an untrained extractive baseline.

\paragraph{CNN-LSTM Extractor}\citep{chen2018fast}
builds hierarchical sentence representations which capture long-range dependencies using a CNN and bi-directional LSTM-RNN layers.
A separate LSTM-based pointer network is applied to the representations to extract summary sentences.

\paragraph{BertExt}\citep{liu2019text}
extends the BERT~\citep{Devlin:19} model with the ability to generate distinct representations for multiple text spans.
Based on those representations the model selects sentences into the extractive summary.

\paragraph{MatchSum}\citep{zhong2020extractive}
formulates extractive summarization as a semantic text matching problem.
Multiple candidate summaries are extracted and embedded as dense vectors using a Siamese-BERT model and matched with the reference text in the semantic space.

\paragraph{BART}\citep{Lewis:19}
uses a denoising autoencoder pre-training strategy designed specifically for NLG tasks.
It has achieved state-of the-art results on many generative tasks, including abstractive text summarization.

\paragraph{T5}\citep{Raffel:19}
approaches transfer learning by unifying multiple NLP tasks into a common text-to-text format.
All tasks are modeled with a large-scale seq-to-seq Transformer architecture in the order of billions of parameters.
The model can be used to generate abstractive summaries using a \textit{summarize:} prefix added to the text.

\paragraph{PEGASUS}\citep{Zhang:19b}
uses a pre-training objective designed for abstractive text summarization which includes masked language modeling and gap sentence generation.
The model achieved state-of-the-art performance on mulitple summarization datasets.


\subsection{Setup}

\paragraph{Modeling}
Computational constraints and input length limits of pre-trained models prevent us from training the baselines on long input sequences.
To circumvent those issues we follow a \textit{generate \& rank} approach for \DatasetName-Chapter and \DatasetName-Book.
We use baseline models fine-tuned on \DatasetName-Paragraph, to \textit{generate} individual summaries for all paragraphs in \DatasetName-Chapter and \DatasetName-Book.
Next, we \textit{rank} the generated summaries based on the model's confidence.
In case of abstractive models we look at the perplexity-level, for extractive models we take the model assigned scores.
As the final chapter- or book-level summary we combine the top-$k$ \textit{ranked} paragraph-summaries, where $k$ is chosen based on summary length statistics in the training set.


\paragraph{Extractive Oracle}
We follow the steps described by~\citet{zhong2020extractive} to generate oracle candidates for the \DatasetName-Paragraph data.
First, we compute a mean ROUGE-\{1,2,L\} score between each sentence in a paragraph and the associated summary.
Next, we select the 5 highest scoring sentences and generate  all combinations of 1, 2, and 3 sentences to serve as candidate oracles.
The final oracle chosen from the set of candidates is the one which maximizes the mean ROUGE-\{1,2,L\} score with the paragraph summary.

\paragraph{Implementation}
Models were implemented in Python using the PyTorch~\citep{Paszke:19} and Huggingface~\citep{Wolf:19} libraries.
Abstractive models were initalized from pretrained checkpoints shared through the Huggingface Model Hub.
Additional details are listed in Appendix~\ref{app:model-checkpoints}.

\paragraph{Training \& Inference}
All models were trained for 10 epochs and evaluated on the validation split at the end of each epoch.
Final model checkpoints were chosen based on the performance of models on the validation data.
Model outputs were decoded using beam search with 5 beams and $n$-gram repetition blocking for $n > 3$~\citep{Paulus:18}.
\begin{table*}[tb]
    \centering
    \resizebox{\linewidth}{!}{%
    \small
    \begin{tabular}{lrrrrrrrrrrrr} 
    \toprule
    & \multicolumn{4}{c}{\DatasetName-Paragraph} & \multicolumn{4}{c}{\DatasetName-Chapter} & \multicolumn{4}{c}{\DatasetName-Book} \\
    \cmidrule(lr){2-5} 
    \cmidrule(lr){6-9}
    \cmidrule(lr){10-13}
    \textbf{Models} & 
    $Flu.$ & $Coh.$ & $Rel.$ & $Fact.$ & 
    $Flu.$ & $Coh.$ & $Rel.$ & $Fact.$ &
    $Flu.$ & $Coh.$ & $Rel.$ & $Fact.$ \\
    \midrule
    BART~\textsubscript{fine-tuned} & 4.24 & 4.15 & 4.03 & 3.96 & 4.16 & 3.96 & - & - & 3.80 & 3.87 & - & - \\
    T5~\textsubscript{fine-tuned} & 4.08 & 4.09 & 3.99 & 3.96 & 4.03 & 4.01 & - & - & 3.93 & 3.91 & - & - \\
    PEGASUS-large~\textsubscript{fine-tuned} & 4.10 & 4.13 & 3.99 & 3.87 & 4.07 & 3.96 & - & - & 3.95 & 3.79 & - & - \\

    \bottomrule
    \end{tabular}
    }%
    \caption{
    Performance of baseline models on the Paragraph, Chapter, and Full-Book subsets of \DatasetName~evaluated by human annotators.
    Judges were asked to assess the fluency (\textit{Flu.}), coherence (\textit{Coh.}), relevance (\textit{Rel.}) and factuality (\textit{Fact.}) of generated summaries.
    Relevance and factuality were not evaluated on the chatper- and book-level due to the length of the source texts.
    }
    \label{tab:human-study}
\end{table*}

\paragraph{Evaluation Metrics}
Models were evaluated using a suite of automatic evaluation metrics included in the SummEval toolkit~\citep{Fabbri:21}.
Lexical overlap between $n$-grams in generated and reference summaries was measured using ROUGE-\{1,2,L\} metrics~\citep{Lin:04}.
Semantic overlap between mentioned summaries was evaluated using BERTScore~\citep{Zhang:20}, which aligns summaries on a token-level based on cosine similarity scores between token embeddings.
We also inspect content overlap between generated summaries and source documents by employing SummaQA~\citep{Scialom:19}, which generates questions based on the input document and next applies a QA system to evaluate how many of those question can be answered using the summary.
Due to the input length limits of SummaQA, the metric was applied individually to paragraphs of chapters and books and next aggregated by averaging to obtain chapter and book-level scores.

\subsection{Automatic Evaluation}
We first evaluate the the baseline models using automatic metrics, with results shown in Table~\ref{tab:rouge-scores}. 

A general trend showing across all evaluated models is low BERTScore values which decrease as reference summaries get longer (from paragraphs to full books).
The metric operates on a $[-1, 1]$ range, and the highest scores, slightly above 0.19, were achieved by the fine-tuned T5 model on a paragraph level.
This suggests that BERTScore might not be a good fit for evaluating highly abstractive, long summaries.
We decided to include it in the evaluation process to highlight this issue for future investigation.

\paragraph{Heuristics}
The performance of the Lead-3 baseline is relatively low, scoring an R-1 of 17.99, 14.32, and 6.50 on the paragraph-, chapter-, and book-level respectively.
The random sentence baseline closely trails Lead-3 across all metrics and data splits.
Both results suggest that data from the literature domain included in \DatasetName~may be less susceptible to layout biases present in other domains, such as newswire.
Extractive oracle scores on paragraph data substantially underperformed those on the chapter and book data.
This could be an artifact of the data pairing procedure where the content of a highly abstractive summary sentences is partially covered by the matched paragraph.

\paragraph{Extractive Models}
The performances of the CNN-LSTM and BertExt models are very similar, with the first model being better on paragraph data, and the second model performing better on chapters and books.
The small performance gap between the two mentioned models is surprising considering that the BERT based model was initialized from a pre-trained checkpoint, while the CNN-LSTM model was trained from scratch.
The MatchSum baseline which reported state-of-the-art performance on news domain datasets~\citep{zhong2020extractive} achieved the best performance on a paragraph level, but underperformed the other models on chapter and book summaries.

\paragraph{Abstractive Models}
We evaluated the performance of abstractive models both in a zero-shot setting and after fine-tuning on the \DatasetName-Paragraph data.
We find that fine-tuning models on the \DatasetName~data leads to consistent improvements across all models and data granularities, with the exception of the BART model on the book-level which performed better in a zero-shot fashion according to the ROUGE metric, and the T5 model on the SQA metrics.
Upon manual inspection of model outputs we noticed that zeroshot models included fragments of dialogues in the summaries which are less likely to be found in reference summaries, this in turn could contribute to the lower evaluation scores of zero-shot baselines. 
The BART model achieved the best performance out of all the baseline models on paragraph- and chapter-level data, while T5 performed best on the book-level.
Despite its state-of-the-art performance on most summarization datasets~\citep{Zhang:19b}, we found PEGASUS to underperform other baseline models, both in the zero-shot and fine-tuned setting.
Examples of generated summaries are shown in Appendix~\ref{app:model-outputs}.

\subsection{Human Evaluation}
To further assess the performance of abstractive baselines, human annotators were hired and asked to evaluate generated summaries across four dimensions: \textit{fluency}, \textit{coherence}, \textit{relevance}, and \textit{factuality}.
Scores were assigned on a Likert scale from 1 to 5, with each example annotated by 3 judges and the scores averaged.
Relevance and factuality were evaluated only on the paragraph-level since both dimensions require an understanding of the source text, which in the case of chapters and books is prohibitively long.
Results are shown in Table~\ref{tab:human-study}.

Similarly to the study using automatic metrics, BART shows strong performance across all dimensions for the paragraph- and chapter-level subsets and slightly underperforms on full books.
The results also show a general decrease in fluency and coherence across all models as the length of the source documents and summaries increases.
This suggests that generating longer passages of fluent and coherent text poses a problem for existing neural models and could be addressed in future work.

\subsection{Discussion}
The \textit{generate \& rank} approach allowed us to overcome the limitations of existing models and apply the baselines to the chapter- and book-level data.
We recognize that generating and scoring sentences independently has drawbacks, namely:
1) generated summaries may lack coherence,
2) content of selected sentences may overlap or be of low significance, which could negatively affect the overall relevance of the summary.
However, the experiments discussed in this section were intended to be groundwork for the introduced task and we leave developing more tailored methods for future work.

The experiment results also show that \DatasetName~poses challenges not only for existing summarization models, but also for evaluation metrics.
The abstractive nature of reference summaries makes lexical overlap measured by ROUGE an inadequate metric for model evaluation~\citep{Fabbri:21}.
Other recently introduced metrics, such as BERTScore and SummaQA, leverage pre-trained neural models, which in turn makes them subject to the same input length limitations as the evaluated summarization models.
While the model-based metrics can be individually applied to chunks of the data and then aggregated, as in the case of SummaQA, such use was not studied by the authors and could affect the reliability of returned scores.
Human-based studies, which are often used to assess dimensions omitted by automatic metrics, are also problematic when conducted with long-form data included in \DatasetName.
For example, assessing factual consistency requires annotators to be familiar with the content of the source document, which in the case of chapters or books could span dozens of pages making such studies unreliable and prohibitively time consuming.

\section{Conclusions}\label{sec:conclusions}
In this work we introduced \DatasetName, a collection of datasets for long-form narrative summarization.
\DatasetName~includes annotations on three levels of granularity of increasing difficulty: paragraph, chapter, and full-book.
Through a quantitative analysis we compare our dataset to existing summarization corpora and show that \DatasetName~sets new challenges for summarization methods.
We trained extractive and abstractive baseline models leveraging state-of-the-art pre-trained architectures to test the performance of current methods on the task of long-narrative summarization and to enable easy comparison with future methods.
We hope our dataset will contribute to the progress made in the field of automatic text summarization.
\section{Limitations}
\paragraph{Data Collection}
Web data is subject to local copyright laws.
For data that is no longer protected by copyright law, we understand the use described within the paper is legally permissible.
For data that is subject to copyright, we understand that such use is allowed under U.S. copyright law's fair use provision.
Depending on how others use this data, the purpose of their use, the jurisdiction they are in, and other factors considered under copyright law, we understand that the decision on whether a specific use case is fair use involves a legal analysis.
It is advisable to obtain legal counsel prior to using such data.
All data described in this work was collected exclusively for the academic purpose of conducting research.
The purpose of using the \DatasetName~data was only for training models and not for public display or any other use.
No data was stored upon completion of the research process.

\paragraph{Data Biases}
The \DatasetName~dataset contains books written or translated into English.
These books are also more than fifty years old and so representative of society in that era.
The various pretrained models we evaluated on our dataset carry biases of the data they were pretrained on.
However, we did not stress test these models for such ethical biases.
We request our users to be aware of these ethical issues in our dataset that might affect their models and evaluations.

\paragraph{Model Evaluation}
In this work, we have used established metrics, such as ROUGE, as well as recently introduced metrics, such as BERTScore and SummaQA, to evaluate the introduced baseline models.
However, such automatic metrics have not been evaluated for use with very long source documents and highly abstractive summaries. 
Thus, might not accurately reflect the true performance of the evaluated models.
Reliable evaluation of highly abstractive summarization models trained on long source documents is an open problem and an area of active research.
Authors using the \DatasetName~data are encouraged to consult appropriate literature whether more robust evaluation methods are available at the time of writing.

\paragraph{Computational Resources}
Considering the length of source documents included in the \DatasetName~dataset, training and evaluation of neural models might require substantial computational resources.

\bibliography{custom}
\bibliographystyle{acl_natbib}
\clearpage
\appendix
\section{Further Implementation Details}\label{app:model-checkpoints}
Model hyperparameters followed the best configurations described by the original authors of the models.
Models were trained for 10 epochs using a batch size of 16.
Many of the baselines presented in this work leveraged pre-trained checkpoints to initialize weights before fine-tuning on the \DatasetName~data.
Table~\ref{tab:model-checkpoints} lists the checkpoints used for each of the baselines and the approximate number of parameters of each model.
Experiments were conducted using 4 NVidia A100 GPUs, all studies described in this paper took an approximate 8 GPU hours.
\begin{table}[!tbhp]
    \centering
    \resizebox{\linewidth}{!}{%
    \small
    \begin{tabular}{lll} 
    \toprule
    \textbf{Model} & \textbf{Checkpoint} & \textbf{Num. parameters} \\
    \midrule
    \multicolumn{2}{c}{Data Alignment Models} \\
    \midrule
    Bi-Encoder (paraphrase) & \texttt{sentence-transformers/paraphrase-distilroberta-base-v1} & 82M \\
    Bi-Encoder (roBERTa) & \texttt{sentence-transformers/stsb-roberta-large} & 355M \\
    Cross-Encoder & \texttt{cross-encoder/stsb-roberta-large} & 355M \\
    \midrule
    \multicolumn{2}{c}{Baseline Models} \\
    \midrule
    MatchSum & \texttt{bert-base-uncased} & 109M \\
    BertExt & \texttt{bert-base-uncased} & 109M \\
    BART & \texttt{facebook/bart-large-xsum} & 406M \\
    T5 & \texttt{t5-large} & 737M \\
    Pegasus & \texttt{google/pegasus-large} & 568M \\
    \bottomrule
    \end{tabular}
    }%
    \caption{
    Hugginface Model Hub checkpoints used to initialize baseline and similarity score models
    }
    \label{tab:model-checkpoints}
\end{table}

\section{Data Alignment Process}\label{app:data-align}
Alignments between book paragraphs and sentences from associated summaries were computed using heuristic methods.
The alignment processed followed two steps described by \citet{Ladhak:20}:
1) similarity scores were computed for all paragraph-sentence pairs,
2) based on the similarity scores paragraph and sentence were aligned using a stable matching algorithm.
Similarity scores between paragraphs and sentences can be computing using different metrics.
In our study, we focused on lexical overlap methods and neural embedding methods.
The first computed a token overlap between paragraphs and sentences using the ROUGE toolkit and treated that as a similarity score.
The second utilized neural networks to embed the text spans into dense vector representations and next computed the similarity score as the cosine distance between such vectors.

To choose the best similarity score metric we conducted a pilot study on a subset of 100 paragraph-sentences pairs sampled from the validation set.
The sampled examples were matched using the procedure described above with different neural models used for embedding the text spans.
The following similarity score methods were considered:
\paragraph{ROUGE-wtd}\citep{Ladhak:20} computes an average of token-weighted ROUGE-\{1,2,L\} scores between the sentence and paragraph texts.
Token weights approximate the saliency of words and are computed as an inverse frequency of word occurrences in the document.

\paragraph{ROUGE-avg}\citep{Ladhak:20} computes an average of (unmodified) ROUGE-\{1,2,L\} scores between the sentence and paragraphs.

\paragraph{BERTScore}\citep{Zhang:20} measures semantic overlap between the words in the sentences and paragraphs.
It aligns words in both text spans by maximizing the cosine similarity between BERT representations of the tokens.

\paragraph{Cross-Encoder}\citep{humeau2019poly} performs self-attention over the sentence and paragraph text passed together through a Transformer network to generate a similarity score between the input pair.

\paragraph{Bi-Encoder}\citep{Reimers:19} uses a Transformer architecture to independently encode the sentence and paragraph texts into a dense vector representation.
The similarity score is calculated using cosine similarity between the sentence and paragraph representations.
We evaluate two checkpoints for the Bi-Encoders as described in Table~\ref{tab:model-checkpoints}.

The quality of data alignments obtained during the pilot study was assessed by human judges hired through the Amazon Mechanical Turk platform.
Workers were hired from English speaking countries and offered a wage of approximately 12 USD per hour.
Annotators were shown paragraphs which were aligned with a shared summary sentence using the different methods.
For each alignment the annotators were asked to label whether the paragraph-sentence pair is \textit{related}, \textit{somewhat related}, or \textit{unrelated}.
Each example was evaluated by three judges, \textit{related} and \textit{somewhat related} labels were merged into a single \textit{positive} label and the majority vote was computed.
Results of the study are presented in Table~\ref{tab:alignment-study} and show the number of times a method was assigned a \textit{positive} label.
The best performing strategy which used a Bi-Encoder fine-tuned on paraphrase detection data.
\begin{table}[h]
    \centering
    \small
    \begin{tabular}{lcc} 
    \toprule
    \textbf{Model} & \textbf{\# selected} \\
    \midrule
    ROUGE-wtd & 74 \\
    ROUGE-avg & 66 \\
    BERTscore & 68 \\
    Cross Encoder & 72 \\
    Bi-Encoder (paraphrase) & \textbf{78} \\
    Bi-Encoder (roBERTa) & 74 \\
    \bottomrule
    \end{tabular}
    
    \caption{
    Number of times an alignment method received a positive label.
    }
    \label{tab:alignment-study}
\end{table}

Using the selected scoring function, paragraph-summary sentence scores were computed between all paragraph-sentence pairs.
Next, this data was input into a stable matching algorithms~\citep{Shapley:62} to obtain the final alignments.
The stable matching procedure creates alignments where no paragraph would prefer to be matched with a different summary sentence to which it is already matched, and no summary sentence would prefer to be matched to another paragraph than it is already matched with.

\section{Alignment Quality}\label{app:align-quality}
The quality of alignments obtained using the process described in Section~\ref{ssec:data-collection} and Appendix~\ref{app:data-align} was also evaluated quantitatively, results are presented in Table~\ref{tab:alignment-scores}
To measure the semantic similarity of source paragraphs and paired summary sentences, the cosine similarity between their embeddings was computed.
To measure lexical overlap between the paragraph-summary pairs ROUGE-1 (R-1), ROUGE-2 (R-2), and ROUGE-L (R-L) scores were computed.
Results are presented in Table~\ref{tab:alignment-scores}.

The cosine similarity of 0.412 indicates strong semantic overlap between the paired sentences and source paragraphs, suggesting high quality pairings.
In comparison, the relatively low lexical overlap of 17.39 R-1 between the mentioned fragments highlights the high abstractiveness of the data.
\begin{table}[th]
    \centering
    \small
    \begin{tabular}{lrrr} 
    \toprule
    Cos. Sim. & $\textrm{R-1}_{f_1}$ & $\textrm{R-2}_{f_1}$ & $\textrm{R-L}_{f_1}$ \\
    \midrule
    0.412 & 17.39 & 3.63 & 11.77  \\
    \bottomrule
    \end{tabular}
    
    \caption{
    Alignment scores between source paragraphs and paired summary sentences.
    Semantic similarity evaluated by means of cosine similarity between the embedded text fragments. 
    Lexical similarity evaluated using ROUGE-1, ROUGE-2, and ROUGE-L f-scores.
    }
    \label{tab:alignment-scores}
\end{table}

\section{Data Creation Pipeline}\label{app:data-pipeline}
\begin{figure*}[tb]
\centering
    \includegraphics[width=0.99\linewidth]{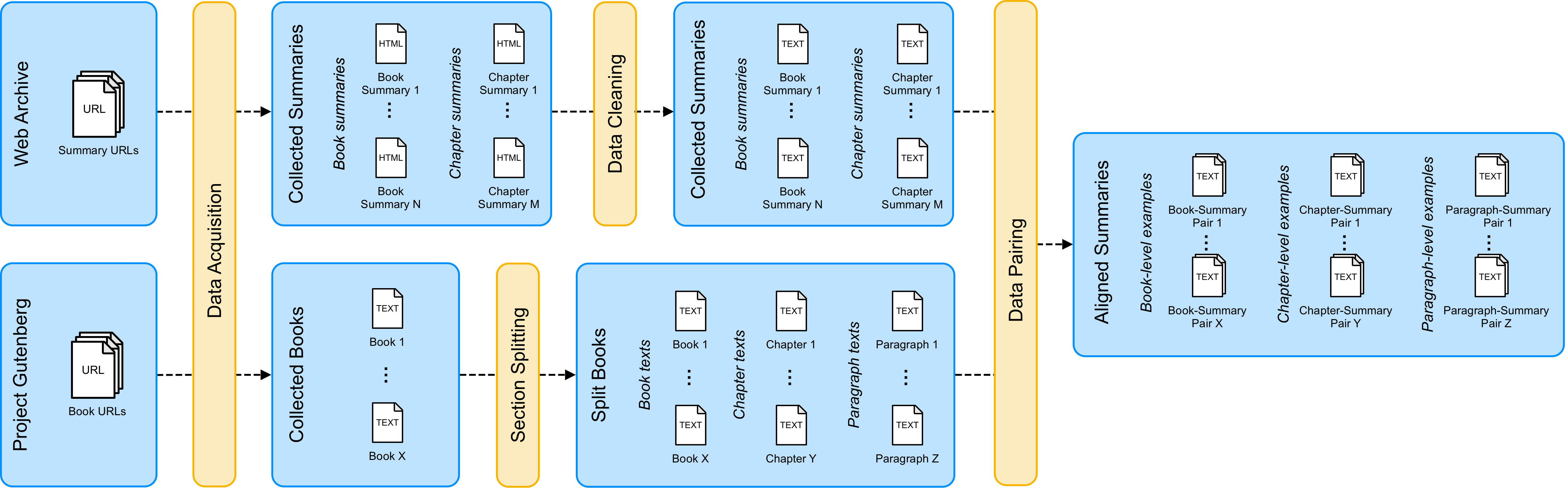}
    \caption{The data collection and pre-processing pipeline used to create the \DatasetName~collection.
    }
    \label{fig:pipeline-figure}
\end{figure*}

The data creation process is visualized in Figure~\ref{fig:pipeline-figure}.

\section{Source examples}
Examples of chapter-level summaries of ”Sense and Sensibility” collected from different sources are shown in Table~\ref{tab:source-examples}.
\begin{table*}[th!]
    \centering
    \small
    \begin{tabular}{p{\linewidth}} 
    \toprule
    \textbf{Text from "Sense and Sensibility", Chapter 1} \\
    \midrule
    \setulcolor{blue}
    \ul{The family of Dashwood had long been settled in Sussex. Their estate was large, and their residence was at Norland Park, in the centre of their property, where, for many generations, they had lived in so respectable a manner as to engage the general good opinion of their surrounding acquaintance.} The late owner of this estate was a single man, who lived to a very advanced age, and \setulcolor{green}\ul{who for many years of his life, had a constant companion and housekeeper in his sister. But her death, which happened ten} (...) 
    \\
    \midrule
    \textbf{Summary from Gradesaver} \\
    \midrule
    \setulcolor{blue}
    \ul{The Dashwood family is introduced; they live at Norland Park, an estate in Sussex, which has been in their family for many years.} \setulcolor{red}\ul{Henry Dashwood has a son by a previous marriage, who is well-off because of his long-deceased mother's fortune; Mr. Dashwood also has three daughters by his present wife}, who are left with very little when he dies and the estate goes to his  (...) 
    \\
    \midrule
    \textbf{Summary from Shmoop} \\
    \midrule
    \setulcolor{blue}
    \ul{We begin with a history of the Dashwood family of Sussex, England}: the head of the family, old Mr. Dashwood, dies and distributes his estate among his surviving relatives: his nephew, \setulcolor{red}\ul{Henry Dashwood, and his children. The children include one son, John, from a first marriage, and three daughters, Elinor, Marianne, and Margaret, from his second.} Even though John and his (...) 
    \\
    \midrule
    \textbf{Summary from Cliffnotes} \\
    \midrule
    \setulcolor{blue}
    \ul{For many years, Henry Dashwood and his family had lived at Norland Park} and cared for its owner, Henry's aged uncle. On the old man's death, Henry inherited the estate. He had always expected that he would be free to leave it, in turn, \setulcolor{red}\ul{to be shared among his wife and three daughters. John, his son by a previous marriage}, was amply provided for. His mother had left him a large  (...) 
    \\
    \midrule
    \textbf{Summary from Sparknotes} \\
    \midrule
    \setulcolor{blue}
    \ul{Old Mr. Dashwood is the owner of a large estate in Sussex called Norland Park}. \setulcolor{green}\ul{Following the death of his sister}, Mr. Dashwood invites his nephew Mr. Henry Dashwood to come live with him at Norland. \setulcolor{red}\ul{The younger Mr. Dashwood brings John Dashwood, his son from a previous marriage, as well as the three daughters born to his present wife.} John Dashwood is grown and (...) 
    \\
    \midrule
    \textbf{Summary from Novelguide} \\
    \midrule
    Sense and Sensibility opens by introducing the Dashwood family, whose fortunes the novel follows. \setulcolor{blue} \ul{The Dashwoods have for many generations owned and occupied the country estate of Norland Park in Sussex, England.} The recent owner, Henry Dashwood, inherited the estate from a Dashwood uncle, referred to as “the old Gentleman.” \setulcolor{red}\ul{Henry Dashwood has a son}, (...) 
    \\
    \midrule
    \textbf{Summary from BarronBooks} \\
    \midrule
    \setulcolor{blue}
    \ul{Mr. Henry Dashwood is leading a comfortable and happy life with his family at Norland Estate}, which belongs to his uncle. He is the rightful heir to the property. However, after his uncle. s death, it is revealed that his son, John Dashwood, and his grandson, Harry, are to inherit the estate. Mr. Henry Dashwood is obviously disappointed. He is concerned about the welfare of his (...) 
    \\
    \bottomrule
    \end{tabular}
    \caption{
    Examples of chapter-level summaries of "Sense and Sensibility" collected from different sources. Text spans underlined with the same color highlight the high semantic and low lexical overlap between the summaries indicating that the summaries are highly abstractive.
    }
    \label{tab:source-examples}
\end{table*}

\section{Human Evaluation UI}\label{app:mturk-ui}
\begin{figure*}[tb]
    \centering
    \includegraphics[width=0.95\linewidth]{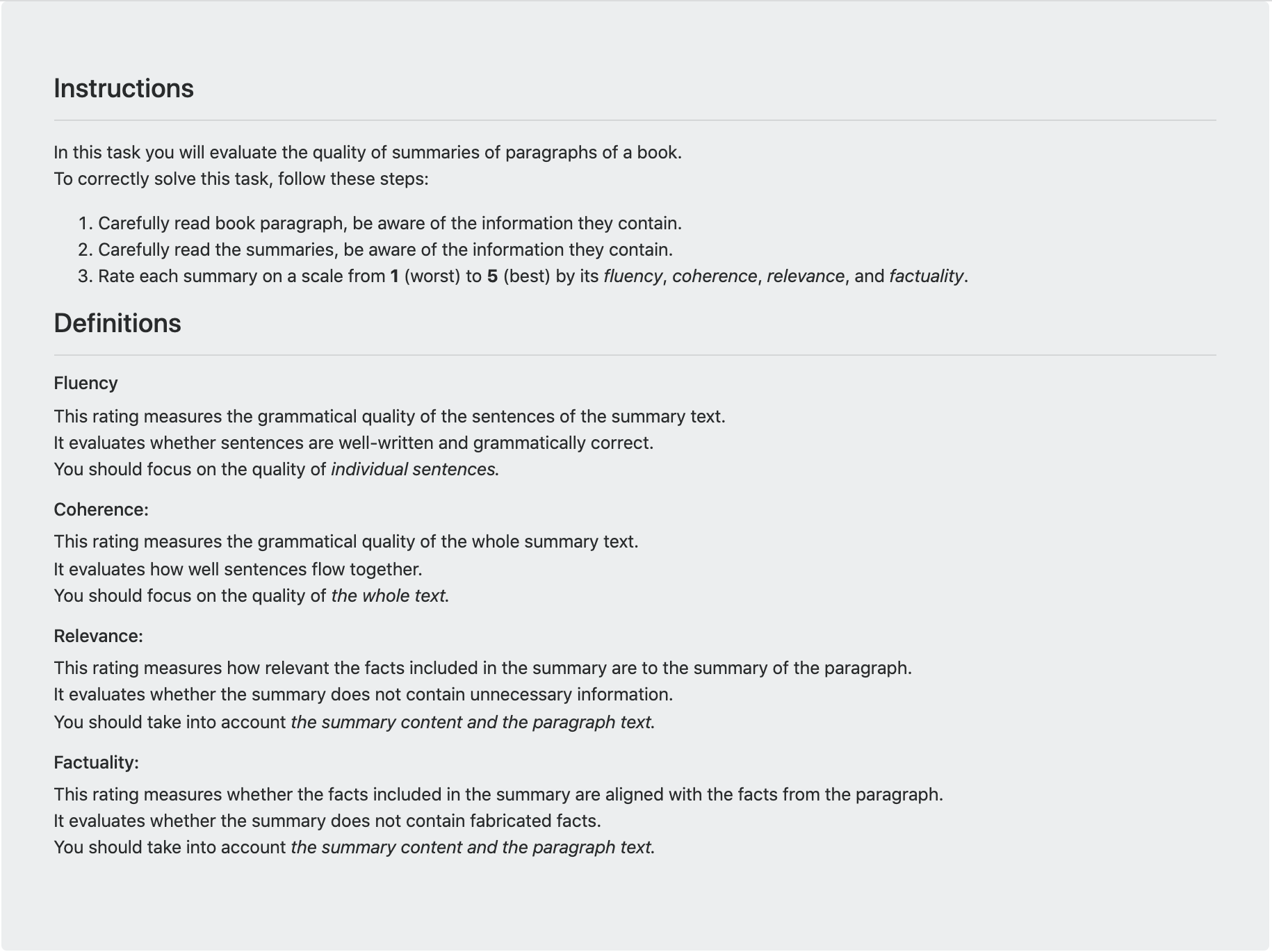}
    \caption{
    Screenshot of the User Interface used to evaluate summaries on the paragraph-level.
    }
    \label{fig:ui-context}
\end{figure*}

\begin{figure*}[tb]
    \centering
    \includegraphics[width=0.95\linewidth]{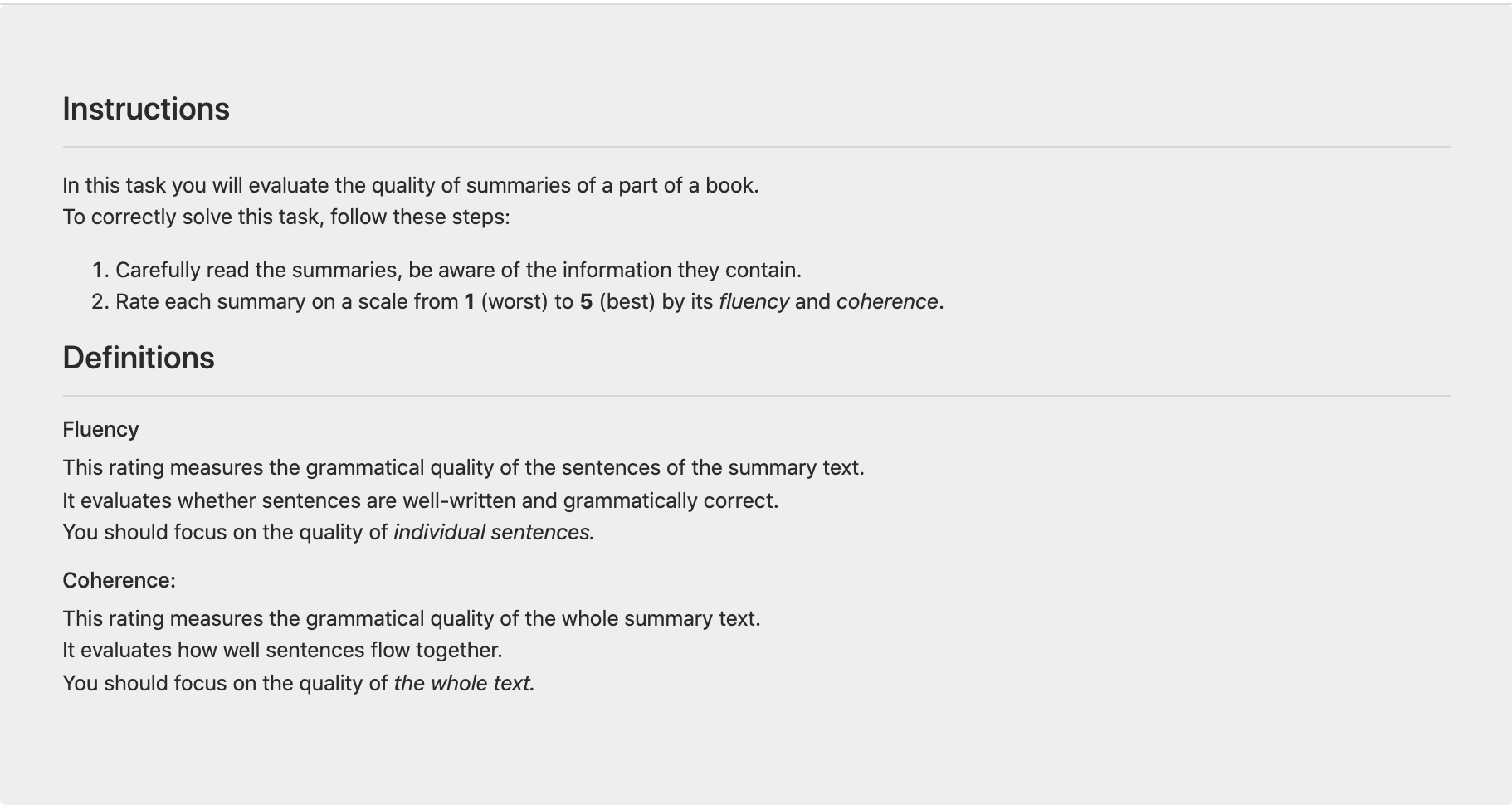}
    \caption{
    Screenshot of the User Interface used to evaluate summaries on the chapter- and book-level.
    }
    \label{fig:ui-nocontext}
\end{figure*}

Screenshots of the user interface, including evaluation instructions, used in thee human studies of abstractive baselines on the paragraph-level are presented in Figure~\ref{fig:ui-context}, and on the chapter- and book-level in Figure~\ref{fig:ui-nocontext}

\section{Model outputs}\label{app:model-outputs}
Example summaries generated on the paragraph-, chapter-, and book-level by the baseline models discussed in our work are presented in Tables~\ref{tab:decoded-summaries-paragraphs}, \ref{tab:decoded-summaries-chapters-part1}, \ref{tab:decoded-summaries-chapters-part2}, \ref{tab:decoded-summaries-books-part1},
\ref{tab:decoded-summaries-books-part2}, \ref{tab:decoded-summaries-books-part3},  \ref{tab:decoded-summaries-books-part4},  \ref{tab:decoded-summaries-books-part5}.
\begin{table*}[!htbp]
    \centering
    \small
    \begin{tabular}{p{\linewidth}} 
    \toprule
    \textbf{Paragraph from ``Sense and Sensibility", Chapter 1} \\
    \midrule
    The family of Dashwood had long been settled in Sussex. Their estate was large, and their residence was at Norland Park, in the centre of their property, where, for many generations, they had lived in so respectable a manner as to engage the general good opinion of their surrounding acquaintance. The late owner of this estate was a single man, who lived to a very advanced age, and who for many years of his life, had a constant companion and housekeeper in his sister. But her death, which happened ten years before his own, produced a great alteration in his home; for to supply her loss, he invited and received into his house the family of his nephew Mr. Henry Dashwood, the legal inheritor of the Norland estate, and the person to whom he intended to bequeath it. In the society of his nephew and niece, and their children, the old Gentleman's days were comfortably spent. His attachment to them all increased. The constant attention of Mr. and Mrs. Henry Dashwood to his wishes, which proceeded not merely from interest, but from goodness of heart, gave him every degree of solid comfort which his age could receive; and the cheerfulness of the children added a relish to his existence. 
    \\
    \midrule
    \textbf{Reference Summary - Paragraph-level (Shmoop)} \\
    \midrule
    We begin with a history of the Dashwood family of Sussex, England: the head of the family, old Mr. Dashwood, dies and distributes his estate among his surviving relatives: his nephew, Henry Dashwood, and his children. Tragically, this arrangement doesn't last long - Henry Dashwood dies, and his estate, including the money he'd recently inherited from his uncle, is re-distributed amongst his wife and children.
    \\
    \midrule
    \textbf{Model Output: CNN-LSTM} \\
    \midrule
    the family of dashwood had long been settled in sussex .
    \\
    \midrule
    \textbf{Model Output: BertExt} \\
    \midrule
    the family of dashwood had long been settled in sussex .
    \\
    \midrule
    \textbf{Model Output: MatchSum} \\
    \midrule
    But her death, which happened ten years before his own, produced a great alteration in his home; for to supply her loss, he invited and received into his house the family of his nephew Mr. Henry Dashwood, the legal inheritor of the Norland estate, and the person to whom he intended to bequeath it.
    \\
    \midrule
    \textbf{Model Output: BART~\textsubscript{fine-tuned}} \\
    \midrule
    The Dashwood family has been settled in Sussex for generations. The late owner of the estate was a single man, who lived to a very advanced age and had a constant companion and housekeeper in his sister. His sister died ten years before his own, and he invited the family of his nephew, Henry Dashwood, into his house.
    \\
    \midrule
    \textbf{Model Output: BART~\textsubscript{zero-shot}} \\
    \midrule
    The story of the death of Henry Dashwood, 1st Earl of Norland, is told in a series of letters written by his son and daughter-in-law.
    \\
     \midrule
    \textbf{Model Output: T5~\textsubscript{fine-tuned}} \\
    \midrule
    The Dashwood family has been settled in Sussex for many generations.
    \\
    \midrule
    \textbf{Model Output: T5~\textsubscript{zero-shot}} \\
    \midrule
    the late owner of the norland estate was a single man, who lived to a very advanced age . to supply her loss, he invited and received into his house the family of his nephew . the old Gentleman's days were comfortably spent with his nephew and niece .
    \\
    \midrule
    \textbf{Model Output: PEGASUS~\textsubscript{fine-tuned}} \\
    \midrule
    The Dashwood family has been settled in Sussex for many generations. Their estate is large, and their residence is at Norland Park, in the center of their property. The late owner of this estate was a single man, who lived to a very advanced age, and who for many years of his life had a constant companion and housekeeper in his sister.
    \\
    \midrule
    \textbf{Model Output: PEGASUS~\textsubscript{zero-shot}} \\
    \midrule
    The late owner of this estate was a single man, who lived to a very advanced age, and who for many years of his life, had a constant companion and housekeeper in his sister.
    \\
    \bottomrule
    \end{tabular}
    \caption{
    Examples of decoded summaries for one paragraph of "Sense and Sensibility, Chapter 1".
    }
    \label{tab:decoded-summaries-paragraphs}
\end{table*}
\begin{table*}[th!]
    \centering
    \tiny
    \begin{tabular}{p{\linewidth}} 
    \toprule
    \textbf{Paragraph from ``Sense and Sensibility", Chapter 1} \\
    \midrule
    The family of Dashwood had long been settled in Sussex. Their estate was large, and their residence was at Norland Park, in the centre of their property, where, for many generations, they had lived in so respectable a manner as to engage the general good opinion of their surrounding acquaintance. The late owner of this estate was a single man, who lived to a very advanced age, and who for many years of his life, had a constant companion and housekeeper in his sister. But her death, which happened (...) 
    \\
    \midrule
    \textbf{Reference Summary - Chapter-level (Shmoop)} \\
    \midrule
    We begin with a history of the Dashwood family of Sussex, England: the head of the family, old Mr. Dashwood, dies and distributes his estate among his surviving relatives: his nephew, Henry Dashwood, and his children. The children include one son, John, from a first marriage, and three daughters, Elinor, Marianne, and Margaret, from his second. Even though John and his (...) 
    \\
    \midrule
    \textbf{Model Output: CNN-LSTM} \\
    \midrule
    the late owner of this estate was a single man, who lived to a very advanced age, and who for many years of his life, had a constant companion and housekeeper in his sister. the family of dashwood had long been settled in sussex. no sooner was his father's funeral over, than mrs. john dashwood, without sending any notice of her intention to her mother-in-law, arrived with her child and their attendants. by a former marriage, mr. henry dashwood had one son: by his present lady, three daughters. so acutely did mrs. dashwood feel this ungracious behaviour, and so earnestly did she despise her daughter-in-law for it, that, on the arrival of the latter, she would have quitted the house for ever, had not the entreaty of her eldest girl induced her first to reflect on the propriety of going, and her own tender love for all her three children determined her afterwards to stay, and for their sakes avoid a breach with their brother. the son, a steady respectable young man, was amply provided for by the fortune of his mother, which had been large, and half of which devolved on him on his coming of age. when he gave his promise to his father, he meditated within himself to increase the fortunes of his sisters by the present of a thousand pounds a-piece. the old gentleman died: his will was read, and like almost every other will, gave as much disappointment as pleasure. elinor saw, with concern, the excess of her sister's sensibility; but by mrs. dashwood it was valued and cherished. he was neither so unjust, nor so ungrateful, as to leave his estate from his nephew;--but he left it to him on such terms as destroyed half the value of the bequest. they encouraged each other now in the violence of their affliction. to him therefore the succession to the norland estate was not so really important as to his sisters; for their fortune, independent of what might arise to them from their father's inheriting that property, could be but small. but her death, which happened ten years before his own, produced a great alteration in his home; for to supply her loss, he invited and received into his house the family of his nephew mr. henry dashwood, the legal inheritor of the norland estate, and the person to whom he intended to bequeath it. he meant not to be unkind, however, and, as a mark of his affection for the three girls, he left them a thousand pounds a-piece. mr. john dashwood had not the strong feelings of the rest of the family; but he was affected by a recommendation of such a nature at such a time, and he promised to do every thing in his power to make them comfortable. she was sensible and clever; but eager in everything: her sorrows, her joys, could have no moderation.
    \\
    \midrule
    \textbf{Model Output: BertExt} \\
    \midrule
    margaret , the other sister , was a good - humored , well - disposed girl ; but as she had already imbibed a good deal of marianne 's romance , without having much of her sense , she did not , at thirteen , bid fair to equal her sisters at a more advanced period of life . marianne 's abilities were , in many respects , quite equal to elinor 's . when he gave his promise to his father , he meditated within himself to increase the fortunes of his sisters by the present of a thousand pounds a-piece . elinor saw , with concern , the excess of her sister 's sensibility ; but by mrs. dashwood it was valued and cherished . by a former marriage , mr. henry dashwood had one son : by his present lady , three daughters . the family of dashwood had long been settled in sussex . no sooner was his father 's funeral over , than mrs. john dashwood , without sending any notice of her intention to her mother - in - law , arrived with her child and their attendants . so acutely did mrs. dashwood feel this ungracious behaviour , and so earnestly did she despise her daughter - in - law for it , that , on the arrival of the latter , she would have quitted the house for ever , had not the entreaty of her eldest girl induced her first to reflect on the propriety of going , and her own tender love for all her three children determined her afterwards to stay , and for their sakes avoid a breach with their brother . the resemblance between her and her mother was strikingly great . he was not an ill - disposed young man , unless to be rather cold hearted and rather selfish is to be ill - disposed : but he was , in general , well respected ; for he conducted himself with propriety in the discharge of his ordinary duties . elinor , this eldest daughter , whose advice was so effectual , possessed a strength of understanding , and coolness of judgment , which qualified her , though only nineteen , to be the counsellor of her mother , and enabled her frequently to counteract , to the advantage of them all , that eagerness of mind in mrs. dashwood which must generally have led to imprudence . but mrs. john dashwood was a strong caricature of himself ; -- more narrow - minded and selfish . had he married a more amiable woman , he might have been made still more respectable than he was : -- he might even have been made amiable himself ; for he was very young when he married , and very fond of his wife . mr. john dashwood had not the strong feelings of the rest of the family ; but he was affected by a recommendation of such a nature at such a time , and he promised to do every thing in his power to make them comfortable . mr. dashwood 's disappointment was , at first , severe ; but his temper was cheerful and sanguine ; and he might reasonably hope to live many years , and by living economically , lay by a considerable sum from the produce of an estate already large , and capable of almost immediate improvement . mrs. john dashwood had never been a favourite with any of her husband 's family ; but she had had no opportunity , till the present , of shewing them with how little attention to the comfort of other people she could act when occasion required it .
    \\
    \midrule
    \textbf{Model Output: MatchSum} \\
    \midrule
    Margaret, the other sister, was a good-humored, well-disposed girl; but as she had already imbibed a good deal of Marianne's romance, without having much of her sense, she did not, at thirteen, bid fair to equal her sisters at a more advanced period of life. No sooner was his father's funeral over, than Mrs. John Dashwood, without sending any notice of her intention to her mother-in-law, arrived with her child and their attendants. When he gave his promise to his father, he meditated within himself to increase the fortunes of his sisters by the present of a thousand pounds a-piece. Mrs. John Dashwood had never been a favourite with any of her husband's family; but she had had no opportunity, till the present, of shewing them with how little attention to the comfort of other people she could act when occasion required it. No one could dispute her right to come; the house was her husband's from the moment of his father's decease; but the indelicacy of her conduct was so much the greater, and to a woman in Mrs. Dashwood's situation, with only common feelings, must have been highly unpleasing;--but in HER mind there was a sense of honor so keen, a generosity so romantic, that any offence of the kind, by whomsoever given or received, was to her a source of immovable disgust. Mr. John Dashwood had not the strong feelings of the rest of the family; but he was affected by a recommendation of such a nature at such a time, and he promised to do every thing in his power to make them comfortable. His son was sent for as soon as his danger was known, and to him Mr. Dashwood recommended, with all the strength and urgency which illness could command, the interest of his mother-in-law and sisters. So acutely did Mrs. Dashwood feel this ungracious behaviour, and so earnestly did she despise her daughter-in-law for it, that, on the arrival of the latter, she would have quitted the house for ever, had not the entreaty of her eldest girl induced her first to reflect on the propriety of going, and her own tender love for all her three children determined her afterwards to stay, and for their sakes avoid a breach with their brother. The prospect of four thousand a-year, in addition to his present income, besides the remaining half of his own mother's fortune, warmed his heart, and made him feel capable of generosity.-- Elinor, this eldest daughter, whose advice was so effectual, possessed a strength of understanding, and coolness of judgment, which qualified her, though only nineteen, to be the counsellor of her mother, and enabled her frequently to counteract, to the advantage of them all, that eagerness of mind in Mrs. Dashwood which must generally have led to imprudence. He was not an ill-disposed young man, unless to be rather cold hearted and rather selfish is to be ill-disposed: but he was, in general, well respected; for he conducted himself with propriety in the discharge of his ordinary duties. Elinor saw, with concern, the excess of her sister's sensibility; but by Mrs. Dashwood it was valued and cherished. But her death, which happened ten years before his own, produced a great alteration in his home; for to supply her loss, he invited and received into his house the family of his nephew Mr. Henry Dashwood, the legal inheritor of the Norland estate, and the person to whom he intended to bequeath it. Mr. Dashwood's disappointment was, at first, severe; but his temper was cheerful and sanguine; and he might reasonably hope to live many years, and by living economically, lay by a considerable sum from the produce of an estate already large, and capable of almost immediate improvement. The constant attention of Mr. and Mrs. Henry Dashwood to his wishes, which proceeded not merely from interest, but from goodness of heart, gave him every degree of solid comfort which his age could receive; and the cheerfulness of the children added a relish to his existence. He meant not to be unkind, however, and, as a mark of his affection for the three girls, he left them a thousand pounds a-piece.
    \\
    \bottomrule
    \end{tabular}
    \caption{
    Examples of decoded summaries of the Chapter 1 of ``Sense and Sensibility", part 1.
    }
    \label{tab:decoded-summaries-chapters-part1}
\end{table*}
    \begin{table*}[th!]
    \centering
    \tiny
    \begin{tabular}{p{\linewidth}} 
    \toprule
    \textbf{Model Output: BART~\textsubscript{fine-tuned}} \\
    \midrule
    The Dashwood family has been settled in Sussex for generations. The late owner of the estate was a single man, who lived to a very advanced age and had a constant companion and housekeeper in his sister. His sister died ten years before his own, and he invited the family of his nephew, Henry Dashwood, into his house. Mr. Dashwood has three daughters, one of whom is his son, and the other two are his sisters. Margaret is a good-humored, well-disposed girl, but as she had already imbibed a good deal of Marianne's romance without having much of her sense, she does not, at thirteen, bid fair to equal her sisters at a more advanced age. Mr. Dashwood dies and leaves his estate to his nephew, a four-year-old boy. Mr. Dashwood's health begins to fail, and he is told that he will not live long. He was not an ill-disposed young man, unless to be rather cold hearted and rather selfish, but he was, in general, well respected. He promises to give his sisters a thousand pounds a year. Mrs. John Dashwood arrives at her husband's funeral without telling her mother-in-law. The eldest daughter, Elinor, is nineteen years old, and she has a "strength of understanding, and coolness of judgment, which qualified her, though only nineteen, to be the counsellor of her mother" . Marianne is described as "sensible and clever; but eager in everything; her sorrows, her joys, could have no moderation." Elinor and her sister-in-law, Mrs. Dashwood, are deeply affected by their mother's death.
    \\
    \midrule
    \textbf{Model Output: BART~\textsubscript{zero-shot}} \\
    \midrule
    The story of the death of Henry Dashwood, 1st Earl of Norland, is told in a series of letters written by his son and daughter-in-law. The death of Henry Dashwood, 1st Viscount Norland, was announced by his son-in-law, the Earl of Wessex. Marianne, the eldest sister, was a good-humored, well-disposed girl; but as she had already imbibed a good deal of Marianne's romance, she did not, at thirteen, bid fair to equal her sisters at a more advanced period of life. The story of Mr. Dashwood, who lived at Norland, is as follows:. When John Dashwood's uncle, who had been one of the richest men in the county, fell seriously ill, it was thought that he might live for many more years. John Dashwood was one of the most respectable men of his time. When George III was a young man, he promised to his father that he would give his sisters a thousand pounds a-year, if he should ever die. No sooner had John Dashwood's father died, than his mother-in-law came to the funeral. So acutely did Mrs. Dashwood feel this ungracious behaviour, and so earnestly did she despise her daughter-in-law for it, that, on the arrival of her eldest daughter, Elinor, to the house, she almost gave up hope of ever seeing her again. Marianne was the second daughter of Elinor, and the only one of her generation to be born in England. Elinor Dashwood and her sister-in-law were in the same state of grief when they arrived at the house, and in the space of a few hours they had become inseparable.
    
    \\
     \midrule
    \textbf{Model Output: T5~\textsubscript{fine-tuned}} \\
    \midrule
    The Dashwood family has been settled in Sussex for many generations. Henry Dashwood has a son and three daughters. Margaret, the other sister, is a good-humored, well-disposed girl. When Mr. Dashwood dies, he leaves his estate to his niece and three daughters. Mr. Dashwood's fortune is only enough to support his family for the next twelve months. Mr. Dashwood is not an ill-disposed young man, but he is, in general, well respected. He decides to give his sisters a thousand pounds each. After his father's funeral, Mrs. Dashwood arrives with her child. Mrs. Dashwood's eldest daughter, Elinor, advises her mother. Marianne's abilities are equal to Elinor's. She is generous, amiable, interesting and everything but prudent. Elinor and Mrs. Dashwood are both deeply afflicted.
    \\
    \midrule
    \textbf{Model Output: T5~\textsubscript{zero-shot}} \\
    \midrule
    the late owner of the norland estate was a single man, who lived to a very advanced age . to supply her loss, he invited and received into his house the family of his nephew . the old Gentleman's days were comfortably spent with his nephew and niece . by a former marriage, mr. dashwood had one son: by his present lady, three daughters . the son was amply provided for by the fortune of his mother, which was large . by his own marriage, which happened soon afterwards, he added to his wealth . but the succession to the norland estate was not so really important as to his sisters . Margaret, the other sister, was a good-humored, well-disposed girl . she did not, at thirteen, bid fair to equal her sisters at a more advanced period of life . but she had already imbibed much of marianne's romance, without having much of her sense . the old gentleman dashwood left his estate to his nephew on such terms as destroyed half the value of the bequest . he left it to his son, and his son's son, a child of four years old, in such a way as to leave to himself no power of providing for those who were most dear to him . the fortune, which had been so tardy in coming, was his only one twelvemonth . ten thousand pounds, including the late legacies, was all that remained for his widow and daughters . his son was sent for as soon as his danger was known, and to him he recommended the interest of his mother-in-law and sisters . had he married a more amiable woman, he might have been made even more respectable than he was . but Mrs. dashwood was a strong caricature of himself;--more narrow-minded and selfish. a young man promised to give his sisters a thousand pounds a-piece . the prospect warmed his heart, and made him feel capable of generosity . he thought of it all day long and for many days successively, and he did not repent . no sooner was her father's funeral over, than Mrs. dashwood arrived . no one could dispute her right to come; the house was her husband's . but the indelicacy of her conduct was so much the greater . the eldest daughter, Elinor, possessed a strength of understanding, and coolness of judgment, which qualified her to be the counsellor of her mother . she had an excellent heart;--her disposition was affectionate, and her feelings were strong; but she knew how to govern them . Marianne's abilities were, in many respects, quite equal to her mother's . she was generous, amiable, interesting: everything but prudent . the resemblance between her and her mother was strikingly great . the agony of grief which overpowered them at first, was voluntarily renewed, was sought for, was created again and again . they gave themselves up wholly to their sorrow, seeking increase of wretchedness in every reflection that could afford it .
    \\
    \midrule
    \textbf{Model Output: PEGAUS~\textsubscript{fine-tuned}} \\
    \midrule
    The Dashwood family has been settled in Sussex for many generations. Their estate is large, and their residence is at Norland Park, in the center of their property. The late owner of this estate was a single man, who lived to a very advanced age, and who for many years of his life had a constant companion and housekeeper in his sister. Mr. Dashwood has one son and three daughters. Margaret, the other sister, was a good-humored, well-disposed girl; but as she had already imbibed a lot of Marianne's romance, without having much of her sense, she did not, at thirteen, bid fair to equal her sisters at a more advanced period of life. Mr. Dashwood dies and leaves his estate to his son and his son's four-year-old son. Mr. Dashwood's son, Mr. John Dashwood, is sent to take care of his mother-in-law and sisters. He is not an ill-disposed young man, unless to be rather cold hearted and rather selfish is to be ill-posed: but he was, in general, well respected; for he conducted himself with propriety in the discharge of his ordinary duties. Had he married a more amiable woman, he might have been made still more respectable than he is, for he was very young when he married and very fond of his wife. He promises to give his sisters a thousand pounds a year. Dashwood's mother-in-law, Mrs. John Dashwood, arrives with her son and their attendants. Mrs. Dashwood's eldest daughter, who is nineteen, advises her mother to stay. She's smart, clever, and generous, but she's not very good at moderation. Mrs. Dashwood and her sister-in-law are both afflicted with grief.
    \\
    \midrule
    \textbf{Model Output: PEGASUS~\textsubscript{zero-shot}} \\
    \midrule
    The late owner of this estate was a single man, who lived to a very advanced age, and who for many years of his life, had a constant companion and housekeeper in his sister. The son, a steady respectable young man, was amply provided for by the fortune of his mother, which had been large, and half of which devolved on him on his coming of age. Margaret, the other sister, was a good-humored, well-disposed girl; but as she had already imbibed a good deal of Marianne\'s romance, without having much of her sense, she did not, at thirteen, bid fair to equal her sisters at a more advanced period of life. Dashwood had wished for it more for the sake of his wife and daughters than for himself or his son;--but to his son, and his son\'s son, a child of four years old, it was secured, in such a way, as to leave to himself no power of providing for those who were most dear to him, and who most needed a provision by any charge on the estate, or by any sale of its valuable woods. John Dashwood had then leisure to consider how much there might prudently be in his power to do for them. John Dashwood was a strong caricature of himself;--more narrow-minded and selfish. The prospect of four thousand a-year, in addition to his present income, besides the remaining half of his own mother\'s fortune, warmed his heart, and made him feel capable of generosity.-- "Yes, he would give them three thousand pounds: it would be liberal and handsome! No one could dispute her right to come; the house was her husband\'s from the moment of his father\'s decease; but the indelicacy of her conduct was so much the greater, and to a woman in Mrs. Elinor, this eldest daughter, whose advice was so effectual, possessed a strength of understanding, and coolness of judgment, which qualified her, though only nineteen, to be the counsellor of her mother, and enabled her frequently to counteract, to the advantage of them all, that eagerness of mind in Mrs. She was generous, amiable, interesting: she was everything but prudent. She could consult with her brother, could receive her sister-in-law on her arrival, and treat her with proper attention; and could strive to rouse her mother to similar exertion, and encourage her to similar forbearance.
    \\
    \bottomrule
    \end{tabular}
    \caption{
    Examples of decoded summaries of the Chapter 1 of ``Sense and Sensibility", part 2.
    }
    \label{tab:decoded-summaries-chapters-part2}
\end{table*}
\begin{table*}[th!]
    \centering
    \tiny
    \begin{tabular}{p{\linewidth}} 
    \toprule
    \textbf{Paragraph from ``Sense and Sensibility", Chapter 1} \\
    \midrule
    The family of Dashwood had long been settled in Sussex. Their estate was large, and their residence was at Norland Park, in the centre of their property, where, for many generations, they had lived in so respectable a manner as to engage the general good opinion of their surrounding acquaintance. The late owner of this estate was a single man, who lived to a very advanced age, and who for many years of his life, had a constant companion and housekeeper in his sister. But her death, which happened (...) 
    \\
    \midrule
    \textbf{Reference Summary - Book-level (Shmoop)} \\
    \midrule
    We begin with a history of the Dashwood family of Sussex, England: the head of the family, old Mr. Dashwood, dies and distributes his estate among his surviving relatives: his nephew, Henry Dashwood, and his children. The children include one son, John, from a first marriage, and three daughters, Elinor, Marianne, and Margaret, from his second. Even though John and his (...) 
    \\
    \midrule
    \textbf{Model Output: CNN-LSTM} \\
    \midrule
    mrs. palmer appeared quite well, and i am commissioned to tell you, that you will certainly see her to-morrow." i am persuaded that mrs. smith suspects his regard for marianne, disapproves of it, (perhaps because she has other views for him,) and on that account is eager to get him away;--and that the business which she sends him off to transact is invented as an excuse to dismiss him. i confess," replied elinor, "that every circumstance except one is in favour of their engagement; but that one is the total silence of both on the subject, and with me it almost outweighs every other." she is not well, she has had a nervous complaint on her for several weeks." she will be more hurt by it, for robert always was her favourite.--she will be more hurt by it, and on the same principle will forgive him much sooner." the parties stood thus: the two mothers, though each really convinced that her own son was the tallest, politely decided in favour of the other. mr. ferrars is to be the man. well, sir," said elinor, who, though pitying him, grew impatient for his departure, "and this is all?" i have entered many a shop to avoid your sight, as the carriage drove by. she fell into violent hysterics immediately, with such screams as reached your brother's ears, as he was sitting in his own dressing-room down stairs, thinking about writing a letter to his steward in the country. affecting that air of playfulness, therefore, which is delightful in a woman one loves, she opened the letter directly, and read its contents. it came from town, and is merely a letter of business." lady elliott was delighted with the thought. he wished her a good morning, and, attended by sir john, left the room. it has been, and is, and probably will always be a heavy misfortune to me, that i have had no necessary business to engage me, no profession to give me employment, or afford me any thing like independence. has not his behaviour to marianne and to all of us, for at least the last fortnight, declared that he loved and considered her as his future wife, and that he felt for us the attachment of the nearest relation? well, then, he is lately dead, marianne, for i am sure there was such a man once, and his name begins with an f." mrs. ferrars, with the utmost liberality, will come forward, and settle on him a thousand a year, if the match takes place. sir john's confidence in his own judgment rose with this animated praise, and he set off directly for the cottage to tell the miss dashwoods of the miss steeles' arrival, and to assure them of their being the sweetest girls in the world. when i got to mr. palmer's, i found charlotte quite in a fuss about the child. sir john had been very urgent with them all to spend the next day at the park. said she, on his taking miss steeles's pocket handkerchief, and throwing it out of window--"he is full of monkey tricks." the man and one of the maids were sent off immediately into devonshire, to prepare the house for their mistress's arrival; for as lady middleton was entirely unknown to mrs. dashwood, she preferred going directly to the cottage to being a visitor at barton park; and she relied so undoubtingly on sir john's description of the house, as to feel no curiosity to examine it herself till she entered it as her own. barton park was about half a mile from the cottage. had it been ten, elinor would have been convinced that at that moment she heard a carriage driving up to the house; and so strong was the persuasion that she did, in spite of the almost impossibility of their being already come, that she moved into the adjoining dressing-closet and opened a window shutter, to be satisfied of the truth. mrs. ferrars was the most unfortunate of women--poor fanny had suffered agonies of sensibility--and he considered the existence of each, under such a blow, with grateful wonder. colonel brandon's partiality for marianne, which had so early been discovered by his friends, now first became perceptible to elinor, when it ceased to be noticed by them. the possibility of colonel brandon's arriving and finding her there, came across her. the nature of her commendation, in the present case, however, happened to be particularly ill-suited to the feelings of two thirds of her auditors, and was so very unexhilarating to edward, that he very soon got up to go away. "as moderate as those of the rest of the world, i believe. what should hinder it?"--but checking her desire, confined herself to this silent ejaculation. neither lady middleton nor mrs. jennings could supply to her the conversation she missed; although the latter was an everlasting talker, and from the first had regarded her with a kindness which ensured her a large share of her discourse. at her time of life, any thing of an illness destroys the bloom for ever! and so saying, she took the screens out of her sister-in-law's hands, to admire them herself as they ought to be admired. she had already repeated her own history to elinor three or four times; and had elinor's memory been equal to her means of improvement, she might have known very early in their acquaintance all the particulars of mr. jennings's last illness, and what he said to his wife a few minutes before he died. he rose up, and walked across the room. " after a visit on colonel brandon's side of only three or four days, the two gentlemen quitted barton together.-- elinor resolving to exert herself, though fearing the sound of her own voice, now said, "is mrs. ferrars at longstaple?" "yes," said he, "they were married last week, and are now at dawlish." "you judged from your knowledge of the colonel's general benevolence; but at least you could not foresee that the opportunity would so very soon occur." but she does every thing well." in this state of her spirits, a letter was delivered to her from the post, which contained a proposal particularly well timed. i could not bear to leave the country in a manner that might lead you, or the rest of the neighbourhood, to suspect any part of what had really passed between mrs. smith and myself--and i resolved therefore on calling at the cottage, in my way to honiton. yet, though smiling within herself at the mistake, she honoured her sister for that blind partiality to edward which produced it. the enclosure of norland common, now carrying on, is a most serious drain. in one moment her imagination placed before her a letter from willoughby, full of tenderness and contrition, explanatory of all that had passed, satisfactory, convincing; and instantly followed by willoughby himself, rushing eagerly into the room to inforce, at her feet, by the eloquence of his eyes, the assurances of his letter. mr. harris was punctual in his second visit;--but
    \\
    \midrule
    \textbf{Model Output: BertExt} \\
    \midrule
    they were interrupted by the entrance of margaret ; and elinor was then at liberty to think over the representations of her mother , to acknowledge the probability of many , and hope for the justice of all . mrs. palmer 's information respecting willoughby was not very material ; but any testimony in his favour , however small , was pleasing to her . elinor was alternately diverted and pained ; but marianne persevered , and saw every night in the brightness of the fire , and every morning in the appearance of the atmosphere , the certain symptoms of approaching frost . " it is not every one , " said elinor , " who has your passion for dead leaves . marianne was astonished to find how much the imagination of her mother and herself had outstripped the truth . i confess , " replied elinor , " that every circumstance except one is in favour of their engagement ; but that one is the total silence of both on the subject , and with me it almost outweighs every other . marianne had now been brought by degrees , so much into the habit of going out every day , that it was become a matter of indifference to her , whether she went or not : and she prepared quietly and mechanically for every evening 's engagement , though without expecting the smallest amusement from any , and very often without knowing , till the last moment , where it was to take her . he saw the necessity of inviting the miss steeles immediately , and his conscience was pacified by the resolution of inviting his sisters another year ; at the same time , however , slyly suspecting that another year would make the invitation needless , by bringing elinor to town as colonel brandon 's wife , and marianne as their visitor . elinor was prevented from making any reply to this civil triumph , by the door 's being thrown open , the servant 's announcing mr. ferrars , and edward 's immediately walking in . " what a pity it is , elinor , " said marianne , " that edward should have no taste for drawing . again he stopped to recover himself ; and elinor spoke her feelings in an exclamation of tender concern , at the fate of his unfortunate friend . the morning was chiefly spent in leaving cards at the houses of mrs. jennings 's acquaintance to inform them of her being in town ; and marianne was all the time busy in observing the direction of the wind , watching the variations of the sky and imagining an alteration in the air . " i do not understand what you mean by interrupting them , " said elinor ; " you were all in the same room together , were not you ? his heart was now open to elinor , all its weaknesses , all its errors confessed , and his first boyish attachment to lucy treated with all the philosophic dignity of twenty - four . " oh , " cried miss steele , looking significantly round at them , " i dare say lucy 's beau is quite as modest and pretty behaved as miss dashwood 's . on being informed of the invitation , mrs. dashwood , persuaded that such an excursion would be productive of much amusement to both her daughters , and perceiving through all her affectionate attention to herself , how much the heart of marianne was in it , would not hear of their declining the offer upon her account ; insisted on their both accepting it directly ; and then began to foresee , with her usual cheerfulness , a variety of advantages that would accrue to them all , from this separation . " you did then , " said elinor , a little softened , " believe yourself at one time attached to her ? " to judge from the colonel 's spirits , however , you have not yet made him equally sanguine . as she said this , she looked earnestly at lucy , hoping to discover something in her countenance ; perhaps the falsehood of the greatest part of what she had been saying ; but lucy 's countenance suffered no change . mrs. dashwood again assured him that no alteration of the kind should be attempted . mrs. palmer , on the contrary , who was strongly endowed by nature with a turn for being uniformly civil and happy , was hardly seated before her admiration of the parlour and every thing in it burst forth . it was lucky , however , for mrs. jennings 's curiosity and elinor 's too , that she would tell any thing without being asked ; for nothing would otherwise have been learnt . but elinor would not give her any encouragement to hope that she should tell her sister . when breakfast was over she walked out by herself , and wandered about the village of allenham , indulging the recollection of past enjoyment and crying over the present reverse for the chief of the morning . elinor could hardly keep her countenance as she assented to the hardship of such an obligation . the shock of colonel brandon 's errand at barton had been much softened to mrs. dashwood by her own previous alarm ; for so great was her uneasiness about marianne , that she had already determined to set out for cleveland on that very day , without waiting for any further intelligence , and had so far settled her journey before his arrival , that the careys were then expected every moment to fetch margaret away , as her mother was unwilling to take her where there might be infection . elinor tried to talk of something else ; but miss steele could not be kept beyond a couple of minutes , from what was uppermost in her mind . he paused for her assent and compassion ; and she forced herself to say , " your expenses both in town and country must certainly be considerable ; but your income is a large one . here she took out her handkerchief ; but elinor did not feel very compassionate . elinor agreed to it all , for she did not think he deserved the compliment of rational opposition . one evening in particular , about a week after colonel brandon left the country , his heart seemed more than usually open to every feeling of attachment to the objects around him ; and on mrs. dashwood 's happening to mention her design of improving the cottage in the spring , he warmly opposed every alteration of a place which affection had established as perfect with him . in a very few weeks from the day which brought sir john middleton 's first letter to norland , every thing was so far settled in their future abode as to enable mrs. dashwood and her daughters to begin their journey . his colour increased ; and with his eyes fixed on the ground he only replied , " you are too good . elinor 's thanks followed this speech with grateful earnestness ; attended too with the assurance of her expecting material advantage to marianne , from the communication of what had passed . lady middleton had sent a very civil message by him , denoting her intention of waiting on mrs. dashwood as soon as she could be assured that her visit would be no inconvenience ; and as this message was answered by an invitation equally polite , her ladyship was introduced to them the next day . marianne 's ideas were still , at intervals , fixed incoherently on her mother , and whenever she mentioned her name , it gave a pang to the heart of poor elinor , who , reproaching herself for having trifled with so many days of illness , and wretched for some immediate relief , fancied that all relief might soon be in vain , that every thing had been delayed too long , and pictured to herself her suffering mother arriving too late to see this darling child , or to see her rational . marianne got up and walked about the room . " four years you have been engaged , " said she with a firm voice . " you have been long acquainted with colonel brandon , have not you ? with a letter in her outstretched hand , and countenance gaily smiling , from the persuasion of bringing comfort , she entered their room , saying , " now , my dear , i bring you something that i am sure will do you good . mrs. dashwood , who could not think a man five years younger than herself , so exceedingly ancient as he appeared to the youthful fancy of her daughter , ventured to clear mrs. jennings from the probability of wishing to throw ridicule on his age . mrs. dashwood looked at elinor with surprise . she determined , after some consideration , that if appearances continued many days longer as unpleasant as they now were , she would represent in the strongest manner to her mother the necessity of some serious enquiry into the affair . mrs. palmer appeared quite well , and i am commissioned to tell you , that you will certainly see her to - morrow . an opportunity was soon to be given to the dashwoods of debating on the rest of the children , as sir john would not leave the house without securing their promise of dining at the park the next day . " have you ever seen the lady ? " the lady then -- miss grey i think you called her -- is very rich ?
    \\
    \bottomrule
    \end{tabular}
    \caption{
    Examples of decoded summaries of the full text of ``Sense and Sensibility", part 1.
    }
    \label{tab:decoded-summaries-books-part1}
    \end{table*}

    \begin{table*}[th!]
    \centering
    \tiny
    \begin{tabular}{p{\linewidth}} 
    \toprule
    \textbf{Model Output: MatchSum} \\
    \midrule
    The morning was chiefly spent in leaving cards at the houses of Mrs. Jennings's acquaintance to inform them of her being in town; and Marianne was all the time busy in observing the direction of the wind, watching the variations of the sky and imagining an alteration in the air. "Oh," cried Miss Steele, looking significantly round at them, "I dare say Lucy's beau is quite as modest and pretty behaved as Miss Dashwood's." An opportunity was soon to be given to the Dashwoods of debating on the rest of the children, as Sir John would not leave the house without securing their promise of dining at the park the next day. "You have been long acquainted with Colonel Brandon, have not you?" "Four years you have been engaged," said she with a firm voice. Elinor's thanks followed this speech with grateful earnestness; attended too with the assurance of her expecting material advantage to Marianne, from the communication of what had passed. With a letter in her outstretched hand, and countenance gaily smiling, from the persuasion of bringing comfort, she entered their room, saying, "Now, my dear, I bring you something that I am sure will do you good." Marianne's lips quivered, and she repeated the word "Selfish?" in a tone that implied--"do you really think him selfish?" In the drawing-room, whither she then repaired, she was soon joined by Mrs. Jennings, with a wine-glass, full of something, in her hand. They were obliged to put an end to such an expectation. Mrs. Palmer, on the contrary, who was strongly endowed by nature with a turn for being uniformly civil and happy, was hardly seated before her admiration of the parlour and every thing in it burst forth. "I am, c." As soon as Elinor had finished it, she performed what she concluded to be its writer's real design, by placing it in the hands of Mrs. Jennings, who read it aloud with many comments of satisfaction and praise. She then went away, walking on tiptoe out of the room, as if she supposed her young friend's affliction could be increased by noise. "To judge from the Colonel's spirits, however, you have not yet made him equally sanguine." With such a confederacy against her--with a knowledge so intimate of his goodness--with a conviction of his fond attachment to herself, which at last, though long after it was observable to everybody else--burst on her--what could she do? Here she took out her handkerchief; but Elinor did not feel very compassionate. Again he stopped to recover himself; and Elinor spoke her feelings in an exclamation of tender concern, at the fate of his unfortunate friend. They were interrupted by the entrance of Margaret; and Elinor was then at liberty to think over the representations of her mother, to acknowledge the probability of many, and hope for the justice of all. "This is beyond every thing!" exclaimed Elinor. "DEAR SIR, Elinor was prevented from making any reply to this civil triumph, by the door's being thrown open, the servant's announcing Mr. Ferrars, and Edward's immediately walking in. Elinor could hardly keep her countenance as she assented to the hardship of such an obligation. "You did then," said Elinor, a little softened, "believe yourself at one time attached to her?" Elinor agreed to it all, for she did not think he deserved the compliment of rational opposition. "M.D." Her second note, which had been written on the morning after the dance at the Middletons', was in these words:-- His heart was now open to Elinor, all its weaknesses, all its errors confessed, and his first boyish attachment to Lucy treated with all the philosophic dignity of twenty-four. "It is not every one," said Elinor, "who has your passion for dead leaves." "And what did the Colonel say?" As she said this, she looked earnestly at Lucy, hoping to discover something in her countenance; perhaps the falsehood of the greatest part of what she had been saying; but Lucy's countenance suffered no change. His colour increased; and with his eyes fixed on the ground he only replied, "You are too good." Marianne was astonished to find how much the imagination of her mother and herself had outstripped the truth. On being informed of the invitation, Mrs. Dashwood, persuaded that such an excursion would be productive of much amusement to both her daughters, and perceiving through all her affectionate attention to herself, how much the heart of Marianne was in it, would not hear of their declining the offer upon HER account; insisted on their both accepting it directly; and then began to foresee, with her usual cheerfulness, a variety of advantages that would accrue to them all, from this separation. "In defence of your protege you can even be saucy." "Have you ever seen the lady?" Marianne got up and walked about the room. Mrs. Dashwood, who could not think a man five years younger than herself, so exceedingly ancient as he appeared to the youthful fancy of her daughter, ventured to clear Mrs. Jennings from the probability of wishing to throw ridicule on his age. Mrs. Palmer's information respecting Willoughby was not very material; but any testimony in his favour, however small, was pleasing to her. He paused for her assent and compassion; and she forced herself to say, "Your expenses both in town and country must certainly be considerable; but your income is a large one." She determined, after some consideration, that if appearances continued many days longer as unpleasant as they now were, she would represent in the strongest manner to her mother the necessity of some serious enquiry into the affair. "M.D." The contents of her last note to him were these:-- One evening in particular, about a week after Colonel Brandon left the country, his heart seemed more than usually open to every feeling of attachment to the objects around him; and on Mrs. Dashwood's happening to mention her design of improving the cottage in the spring, he warmly opposed every alteration of a place which affection had established as perfect with him. Elinor was alternately diverted and pained; but Marianne persevered, and saw every night in the brightness of the fire, and every morning in the appearance of the atmosphere, the certain symptoms of approaching frost. In a very few weeks from the day which brought Sir John Middleton's first letter to Norland, every thing was so far settled in their future abode as to enable Mrs. Dashwood and her daughters to begin their journey. Elinor was obliged to turn from her, in the middle of her story, to receive the rest of the party; Lady Middleton introduced the two strangers; Mrs. Dashwood and Margaret came down stairs at the same time, and they all sat down to look at one another, while Mrs. Jennings continued her story as she walked through the passage into the parlour, attended by Sir John. "What! is Fanny ill?" Elinor could NOT remember it;--but her mother, without waiting for her assent, continued, Lucy was all exultation on being so honorably distinguished; and Miss Steele wanted only to be teazed about Dr. Davies to be perfectly happy.
    \\
    \midrule
    \textbf{Model Output: BART~\textsubscript{fine-tuned}} \\
    \midrule
    Margaret is a good-humored, well-disposed girl, but as she had already imbibed a good deal of Marianne\'s romance, without having much of her sense, she does not, at thirteen, bid fair to equal her sisters at a more advanced age. A gentleman carrying a gun was passing up the hill and within a few yards of Marianne when her accident happened. The gentleman offered his services; and perceiving that her modesty declined what her situation rendered necessary, took her up in his arms without farther delay, and carried her down the hill. Sir John tells them that Mr. Willoughby has no property of his own in the country, but that he does have a pretty little estate in Somersetshire. Marianne says that she hates "every common-place phrase by which wit is intended; and 'setting one's cap at a man,' or 'making a conquest'. Elinor tells Marianne that she has been "too much at ease, too happy, too frank" and that she should have been "spiritless, dull, and deceitful." Marianne tells Willoughby that he is a "very respectable man, who has more money than he can spend, more time than he knows how to employ, and two new coats every year." Willoughby says that he dislikes Colonel Brandon for three reasons: he threatened him with rain when he wanted it to be fine, he has found fault with the hanging of his curricle, and he cannot persuade him to buy his brown horse. He tells Elinor that he once knew a lady who thought and judged like her, but who "from an inforced change -- from a series of unfortunate circumstances" . The next day, Marianne tells Elinor that Willoughby has given her a horse. The next day, Margaret tells Elinor that she knows that Marianne is engaged to Mr. Willoughby. Margaret tells Elinor that she saw him cut off a lock of Marianne's hair and put it in his pocketbook. By ten o'clock the whole party was assembled at the park, where they were to breakfast. The morning was rather favourable, though it had rained all night. Elinor hears Willoughby tell Marianne that there are some people who cannot bear a party of pleasure. He says that if he were rich enough, he would "pull Combe down and build it up again in the exact plan of this cottage." He tells her that his mother was "clogged with the payment of three to old superannuated servants by his father's will, and it is amazing how disagreeable she found it." Marianne says that money can only give happiness where there is nothing else to give it. The next morning, Mrs. Dashwood tells him that he would be happier if he had a profession to engage his time and give an interest to his plans and actions. Mrs. Palmer, who is "strongly endowed by nature with a turn for being uniformly civil and happy, was hardly seated before her admiration of the parlor and every thing in it burst forth." Elinor does not think that he is "so genuinely and unaffectedly ill-natured or ill-bred as he wished to appear." He tells her that Mr. Palmer is "exceedingly pleased with you and your sisters and you can't think how disappointed he will be if you don't come to Cleveland." The Palmers return to Cleveland the next day, and the two families are left to entertain each other. She tells him that she would rather be thought impertinently curious than be thought so by a person whose good opinion is so well worth having as yours. Elinor is shocked to learn that they have been engaged for four years. Elinor tells Lucy that Edward's love for her has been "pretty well put to the test, by our long, very long absence since we were first engaged, and it has stood the trial so well, that I should be unpardonable to doubt it now." He was not an ill-disposed young man, unless to be rather cold hearted and rather selfish, but he was, in general, well respected. Mrs. Jennings says that it is odd that a man should use such a pretty girl so ill, but when there is plenty of money on one side, and next to none on the other, they care no more about such things. He tells her that Delaford is a "nice place, full of comforts and conveniences; quite shut in with great garden walls that are covered with the best fruit trees in the country. Marianne is astonished to find how much the imagination of her mother and herself had outstripped the truth. He asks Elinor to walk with him to Conduit Street, where he meets Sir John and Lady Middleton. Mrs. Ferrars is a "little, thin woman, upright, even to formality, in her figure, and serious even to sourness" . The dinner is a grand one, the servants are numerous, and every thing bespoke the Mistress's inclination for show, and the Master's ability to support it. She says that Edward has the most delicate conscience in the world, and that he is "the most fearful of giving pain, of wounding expectation, and the most incapable of being selfish of any body I ever saw." Mr. John Dashwood tells his mother how sorry he is that she has taken a house at such a distance from Norland as to prevent his being of any service to her in removing her furniture. Lady Middleton does not like Marianne and Elinor because they are "too much sense to be desirable companions to the former; and by the latter they were considered with a jealous eye." Ferrars says that Lucy is "a very deserving young woman, but in the present case you know, the connection must be impossible. And to have entered into a secret engagement with a young man under her uncle's care, the son of a woman especially of such a very large fortune as Mrs. Ferrars, is perhaps, altogether a little extraordinary." The first part of their journey is tedious and unpleasant, but as they draw towards the end of it, their interest in the appearance of the country which they were to inhabit overcame their dejection, and the view of Barton Valley gives them cheerfulness. The eldest daughter, Elinor, is nineteen years old, and she has a "strength of understanding, and coolness of judgment, which qualified her, though only nineteen, to be the counsellor of her mother" . The next morning, Elinor receives a letter from Lucy. He tells her that he has heard Mrs. Ferrars say that it would have been "the least evil of the two, and she would be glad to compound now for nothing worse." Elinor finds Mr. Palmer to be "perfectly the gentleman in his behavior to all his visitors, and only occasionally rude to his wife and her mother" . The comfort of such a friend at that moment as Colonel Brandon -- or such a companion for her mother -- how gratefully was it felt -- a companion whose judgment would guide, whose attendance must relieve, and whose friendship might soothe her. Sir John is a "blessing to all the juvenile part of the neighborhood, for in summer he was for ever forming parties to eat cold ham and chicken out of doors, and in winter his private balls were numerous enough for any young lady who was not suffering under the unsatiable appetite of fifteen." Elinor thinks about the "irreparable injury which too early an independence and its consequent habits of idleness, dissipation, and luxury, had made in the mind, the character, the happiness, of a man who, to every advantage of person and talents, united a disposition naturally open and honest, and a feeling, affectionate temper." Mrs. Jennings, Lady Middleton's mother, is a good-humored, merry, fat, elderly woman who talks a lot. When he arrives, he is "not only in the rapturous profession of the lover, but, in the reality of reason and truth, one of the happiest of men." Marianne Dashwood is born to an extraordinary fate. She was born to discover the falsehood of her own opinions, and to counteract, by her conduct, her most favourite maxims. Marianne says that a woman of seven and twenty can never hope to feel or inspire affection again, and if her home be uncomfortable, or her fortune small, she might bring herself to submit to the offices of a nurse for the provision and security of a wife.
    \\
    \bottomrule
    \end{tabular}
    \caption{
    Examples of decoded summaries of the full text of ``Sense and Sensibility", part 2.
    }
    \label{tab:decoded-summaries-books-part2}
    \end{table*}
    \begin{table*}[th!]
    \centering
    \tiny
    \begin{tabular}{p{\linewidth}} 
    \toprule
    \textbf{Model Output: BART~\textsubscript{zero-shot}} \\
    \midrule
    A continuance in a place where everything reminded her of former delight, was exactly what suited her mind. "I do not believe," said Mrs. Dashwood, with a good humoured smile, "that Mr. Willoughby will be incommoded by the attempts of either of my daughters towards what you call CATCHING him. Mrs. John Dashwood did not at all approve of what her husband intended to do for his half sisters. Mrs. Dashwood entered into all their feelings with a warmth which left her no inclination for checking this excessive display of them. "If the impertinent remarks of Mrs Jennings are to be the proof of impropriety in conduct, we are all offending every moment of our lives. "If they were one day to be your own, Marianne, you would not be justified in what you have done." They thanked her; but were obliged to resist all her entreaties. "Oh, don\'t be so sly before us," said Mrs. Palmer, "for we know all about it; and I admire your taste very much, for I think he is extremely handsome. "But I do assure you it was so, for all that, and I will tell you how it happened. "I was afraid you would think I was taking a great liberty with you," said she, "but I have not known you long to be sure, personally at least, but I have known you and all your family by description a great while; and as soon as I saw you, I felt almost as if you was an old acquaintance. "Oh, Lord! I am sure your mother can spare you very well, and I beg you will favour me with your company, for I\'ve quite set my heart upon it. "Pray, pray be composed," cried Elinor, "and do not betray what you feel to every body present." Before the house-maid had lit their fire the next day, or the sun gained any power over a cold, gloomy morning in January, Marianne was kneeling against one of the window-seats, and writing as fast as a continual flow of tears would permit her. "For shame, for shame, Miss Dashwood!" "Well then, another day or two, perhaps; but I cannot stay here long, I cannot endure the questions and remarks of all these people." To give the feelings or the language of Mrs. Dashwood on receiving and answering Elinor\'s letter would be only to give a repetition of what her daughters had already felt and said; of a disappointment hardly less painful than Marianne\'s, and an indignation even greater. The rest of Mrs. Palmer\'s sympathy was shewn in procuring all the particulars in her power of the approaching marriage, and communicating them to Elinor. Elinor had always thought it would be more prudent for them to settle at some distance from Norland, than immediately amongst their present acquaintance. "Poor Marianne!" said her brother to Colonel Brandon, as soon as he could secure his attention. All these jealousies and discontents, however, were so totally unsuspected by Mrs. Jennings. I come now to the relation of a misfortune, which about this time befell Mrs. Jennings. Elinor agreed to it all, for she did not think he deserved the compliment of rational opposition. Here Mrs. Jennings ceased, and as Elinor had had time enough to collect her thoughts, she was able to give such observations, as the subject might naturally be supposed to produce. "Very well indeed!--how prettily she writes!--aye, that was quite proper to let him be off if he would. Perhaps Mrs. Jennings was in hopes, by this vigorous sketch of their future ennui, to provoke him to make an offer, which might give himself an escape from it;--and if so, she had soon afterwards good reason to think her object gained; for, on Elinor\'s moving to the window to take more expeditiously the dimensions of a print, which she was going to copy for her friend. "I never was better pleased in my life, and I wish you joy of it with all my heart." Elinor had just been congratulating herself, in the midst of her perplexity, that however difficult it might be to express herself properly by letter, it was at least preferable to giving the information by word of mouth, when her visitor entered the room. What Edward felt, as he could not say it himself, it cannot be expected that any one else should say for him. Elinor, while she waited in silence and immovable gravity, could not restrain her eyes from being fixed on him with a look that spoke all the contempt it excited. The comfort of such a friend at that moment as Colonel Brandon--or such a companion for her mother,--how gratefully was it felt!--a companion whose judgment would guide, whose attendance must relieve, and whose friendship might soothe her!--as far as the shock of the summons COULD be lessened to her, his presence, his manners, his assistance, would lessen it. "Pray be quick, sir,"--said Elinor, impatiently;--"I have no time to spare." "I do not know," said he, after a pause of expectation on her side, and thoughtfulness on his own, "how YOU may have accounted for my behaviour to your sister, or what diabolical motive you may have imputed to me." "To have resisted such attractions, to have withstood such tenderness!--Is there a man on earth who could have done it?" "Yes, but I had only the credit of servilely copying such sentences as I was ashamed to put my name to. As soon as Mrs. Dashwood had recovered herself, to see Marianne was her first desire; and in two minutes she was with her beloved child, rendered dearer to her than ever by absence, unhappiness, and danger. "His character, however," answered Elinor, "does not rest on ONE act of kindness, to which his affection for Marianne, were humanity out of the case, would have prompted him. "Oh! my love, I could not then talk of hope to him or to myself. "As for regret," said Marianne, "I have done with that, as far as he is concerned." Mrs. Dashwood did not hear unmoved the vindication of her former favourite. "You consider the matter," said Elinor, "as a good mind and a sound understanding must consider it; and I dare say you perceive, not only in this, but in many other circumstances, reason enough to be convinced that your marriage must have involved you in many certain troubles and disappointments, in which you would have been poorly supported by an affection, on his side, much less certain. "Who told you that Mr Ferrars was married, Thomas?" Mrs. Dashwood, who could not think a man five years younger than herself, ventured to clear Mrs. Jennings from the probability of wishing to throw ridicule on his age. How soon he had walked himself into the proper resolution, however, in what manner he expressed himself, and how he was received, need not be told. The change which a few hours had wrought in the minds and the happiness of the Dashwoods, was such--so great--as promised them all, the satisfaction of a sleepless night. A three weeks\' residence at Delaford, where, in his evening hours at least, he had little to do but to calculate the disproportion between thirty-six and seventeen, brought him to Barton in a temper of mind which needed all the improvement in Marianne\'s looks, and all the encouragement of her mother\'s language, to make it cheerful. The whole of Lucy\'s behaviour in the affair, and the prosperity which crowned it, may be held forth as a most encouraging instance of what an earnest, an unceasing attention to self-interest will do in securing every advantage of fortune, with no other sacrifice than that of time and conscience. "A woman of seven and twenty," said Marianne, "can never hope to feel or inspire affection again, and if her home be uncomfortable, or her fortune small, I can suppose that she might bring herself to submit to the offices of a nurse."
    \\
    \midrule
    \textbf{Model Output: T5~\textsubscript{fine-tuned}} \\
    \midrule
    Margaret, the other sister, is a good-humored, well-disposed girl, but she does not, at thirteen, bid fair to equal her sisters at a more advanced period of life. The Dashwoods are now settled at Barton with tolerable comfort to themselves. His beauty and grace are instantly the theme of general admiration, and the laugh which his gallantry raised against Marianne received particular spirit from his exterior attractions. Willoughby tells her that he has three reasons for disliking Colonel Brandon: he threatened her with rain when she wanted it fine, he found fault with the hanging of her curricula, and he cannot persuade him to buy my brown mare. Mrs. Dashwood enters into all their feelings with a warmth which left her no inclination for checking this excessive display of them. Elinor finds Colonel Brandon the only person who can claim the respect of abilities, excite the interest of friendship, or give pleasure as a companion. He says that he once knew a lady who thought and judged like Elinor, but who from an inforced change--from a series of unfortunate circumstances. The next morning, Marianne tells her sister that Willoughby has given her a horse. The sudden termination of Colonel Brandon's visit at the park fills the mind and raises the wonder of Mrs. Jennings for two or three days. One evening, about a week after Colonel Brandon leaves the country, his heart seems more than usual open to every feeling of attachment to the objects around him. Mrs. Dashwood says that she would be puzzled to spend so large a fortune myself if my children were all to be rich without my help. Sir John invites the Dashwoods to dine with him the next day. Mrs. Dashwood tells Edward that he would be happier if he had a profession to engage his time and give an interest to his plans and actions. Mrs. Palmer asks the Dashwoods to come to Cleveland for Christmas. Mrs. Palmer's information about Willoughby is not very material, but any testimony in his favor is pleasing to her. The Palmers return to Cleveland the next day, and the two families at Barton are left to entertain each other. Mrs. Palmer and two elderly ladies of Mrs. Jennings's intimate acquaintance dine with them, and Elinor is obliged to assist in making a whist table for the others. The morning is spent in leaving cards at the houses of Mrs. Jennings' acquaintances to inform them of her being in town. The Miss Dashwoods have no greater reason to be dissatisfied with Mrs. Jennings's style of living, and set of acquaintances, than with her behavior to herself. Mr. and Mrs. Palmer are of the party; from the former, whom they had not seen before since their arrival in town, they received no mark of recognition on their entrance. Elinor is left with a melancholy impression of Colonel Brandon's unhappiness, and is prevented even from wishing it removed by her anxiety for the very event which will confirm it. Marianne is astonished to find how much the imagination of her mother and herself had outstrip the truth. Mrs. Dashwood decides that it would be better for Marianne to be anywhere, at that time, than at Barton, where everything would be bringing back the past in the strongest and most afflicting manner. Mrs. Jennings, who knows nothing of all this, begins to think that they will not be married until Michaelmas, and by the end of a week that it would not be a match at all. Marianne is spared from the troublesome feelings of contempt and resentment, on this impertinent examination of their features, and on the puppyism of his manner in deciding on the different horrors of the toothpick-cases presented to him. After his father's funeral, Mrs. Dashwood arrives with her child. Mrs. Ferrars is a "little, thin woman, upright, even to formality" . The dinner is a grand one, the servants are numerous, and everything bespoke the Mistress' inclination for show and the Master's ability to support it. Miss Steele is the least discomposed by their presence, and if they had only given her a full and minute account of the whole affair between Marianne and Mr. Willoughby, she would have thought herself amply rewarded for the sacrifice of the best place by the fire after dinner. Marianne has become so accustomed to going out every day that it has become a matter of indifference to her whether she goes or not. Mrs. Jennings praises Edward's conduct, but only Elinor and Marianne understand its true merit. The next morning, Elinor receives a letter from Lucy. Marianne's pianoforte is unpacked and properly disposed of; Elinor's drawings are affixed to the walls of their sitting room. Marianne's thoughts are still fixed incoherently on her mother, and whenever she mentions her name, it gives a pang to the heart of poor Elinor, who, reproaching herself for having trifled with so many days of illness, fancied that all relief might soon be in vain, that everything had been delayed too long, and pictured to herself her suffering mother arriving too late to see this darling child. At seven o'clock, Elinor joins Mrs. Jennings in the drawing room for tea. The arrival of a new family in the country is always a joy to him. The Miss Dashwoods are young, pretty, and unaffected. Elinor's thoughts are fixed on the irreparable injury which too early an independence and its consequent habits of idleness, dissipation, and luxury had made in the mind, the character, the happiness, of a man who, to every advantage of person and talents, united a disposition naturally open and honest, and a feeling, affectionate temper. Extravagance and vanity had made him cold-hearted and selfish. Mrs. Jennings, Lady Middleton's mother, is full of jokes and laughter, and before dinner was over had said many witty things on the subject of lovers and husbands. Willoughby's influence over her mind is heightened by circumstances which ought not to have weight; by that person of uncommon attraction, that open, affectionate, and lively manner which no merit to possess; and by that still ardent love for Marianne, which she is not even innocent to indulge. Colonel Brandon, the friend of Sir John, seems no more adapted by resemblance of manner to be his friend than Lady Middleton was to be Sir John's wife, or Mrs. Jennings to be her mother. The Dashwoods are two days on the road, and Marianne bore her journey on both, without essential fatigue. Elinor asks if Mrs. Ferrars is at Longstaple. Edward goes to Barton to ask Elinor to marry him. His heart is now open to Elinor, all its weaknesses confessed, and his first boyish attachment to Lucy treated with the philosophic dignity of twenty-four. The first month after their marriage is spent with their friend at the Mansion-house, from which they could superintend the progress of the Parsonage. She was born to discover the falsehood of her own opinions, and to counteract, by her conduct, her most favourite maxims. Marianne says that a woman of seven and twenty can never hope to feel or inspire affection again, and if her home be uncomfortable, or her fortune small, she might bring herself to submit to the offices of a nurse, for the sake of the provision of security.
    \\
    \bottomrule
    \end{tabular}
    \caption{
    Examples of decoded summaries of the full text of ``Sense and Sensibility", part 3.
    }
    \label{tab:decoded-summaries-books-part3}
    \end{table*}

    \begin{table*}[th!]
    \centering
    \tiny
    \begin{tabular}{p{\linewidth}} 
    \toprule
    \textbf{Model Output: T5~\textsubscript{zero-shot}} \\
    \midrule
    a gentleman carrying a gun, with two pointers playing round him, was passing up the hill and within a few yards of Marianne, when her accident happened . he put down his gun and ran to her assistance; she had raised herself from the ground, but her foot had been twisted in her fall, and she scarcely able to stand . perceiving that her modesty declined what her situation rendered necessary, he took her up in his arms without further delay, and carried her down the hill . "in defence of your protege you can even be saucy" "My protege, as you call him, is a sensible man," says marianne . "he has seen a great deal of the world; has been abroad, has read, and has a thinking mind." marianne abhorred all concealment where no real disgrace could attend unreserved; and to aim at the restraint of sentiments which were not in themselves illaudable, appeared to her an unnecessary effort . and their behaviour at all times, was an illustration of their opinions . "there are inconveniences attending such feelings as marianne's," said Elinor . "her systems have all the unfortunate tendency of setting propriety at nought" "a better acquaintance of the world is what i look forward to as her greatest possible advantage" "i have not known him long indeed, but I am much better acquainted with him, than I am with any other creature in the world, except yourself and mama" "of John I know very little, though we have lived together for years; but of Willoughby my judgment has long been formed." "i am particularly sorry, ma'am," said he, addressing lady middleton . "my own loss is great," he continued, "in being obliged to leave so agreeable a party" "what can you have to do in town at this time of year?" cried Mrs. jennings . "if the impertinent remarks of Mrs. jennings are to be the proof of impropriety in conduct, we are all offending every moment of our lives" "i am not sensible of having done anything wrong in walking over Mrs. Smith's grounds, or in seeing her house" "if they were one day to be your own, marianne, you would not be justified in what you have done." "nothing of the kind will be done; for my mother will never have money" "i am heartily glad of it," he cried . "may she always be poor, if she can employ her riches no better." "if you were to see them at the altar, you would suppose they were going to be married. ungracious girl!" "nothing in my opinion has ever passed to justify doubt; no secrecy has been attempted; all has been uniformly open and unreserved." marianne would have thought herself very inexcusable had she been able to sleep at all the first night after parting with willoughby . but the feelings which made such composure a disgrace, left her in no danger of incurring it . she was awake the whole night, and she wept the greatest part of it; giving pain every moment to her mother and sisters . "i believe you are right, my love; it will be better that there should be no annuity in the case; whatever I may give them occasionally will be of far greater assistance than a yearly allowance" "a present of fifty pounds, now and then, will prevent their ever being distressed for money, and will, I think, be amply discharging my promise to my father." "money can only give happiness where there is nothing else to give it," said marianne . "beyond a competence, it can afford no real satisfaction, as far as mere self is concerned." the fortune, which had been so tardy in coming, was his only one twelvemonth . ten thousand pounds, including the late legacies, was all that remained for his widow and daughters . his son was sent for as soon as his danger was known, and to him he recommended the interest of his mother-in-law and sisters . "to give such intelligence to a person who could not be interested in it, even if it were true, is not what i should expect Colonel Brandon to do." "you have been long acquainted with colonel Brandon, have not you?" "did not colonel Brandon know of sir John's proposal to your mother before it was made? had he never owned his affection to yourself?" "he was four years with my uncle, who lives at longstaple, near Plymouth" "i was very unwilling to enter into it without the knowledge and approbation of his mother" "though you do not know him so well, you must have seen enough to be sensible he is very capable of making a woman sincerely attached to him." "i begged him to exert himself for fear you should suspect what was the matter" "i heard from him just before I left Exeter;" taking a letter from her pocket . "he was tired, I dare say, for he had just filled the sheet to me as full as possible." "it would be madness to marry upon that, though for my own part, i could give up every prospect of more without a sigh." "i love him too well to be the selfish means of robbing him, perhaps, of all that his mother might give him if he married to please her." "it will not be in my power to accept their invitation if they do." "i have a notion," said sir John, "that miss marianne would not object to such a scheme, if her elder sister would come into it." "i would advise you two, to set off for town, when you are tired of barton, without saying a word to miss dashwood about it." the middletons were to follow in about a week . the miss steeles kept their station at the park . if she DID, the letter was written and sent away with a privacy which eluded all her watchfulness to ascertain the fact . but while she saw Marianne in spirits, she could not be very uncomfortable herself . "i am much concerned to find there was anything in my behaviour last night that did not meet your approbation, and I entreat your forgiveness of what I can assure you to have been perfectly unintentional." "my esteem for your whole family is very sincere; but if I have been so unfortunate as to give rise to a belief of more than I felt, or meant, I shall reproach myself for not having been more guarded." "i could rather believe every creature . acquaintance leagued together to ruin me in his opinion, than . believe his nature capable of such cruelty" "is there a creature in the world whom i would not rather suspect of evil than willoughby?" "when a young man, be who he will, comes and makes love to a pretty girl, and promises marriage, he has no business to fly off from his word" "why don't he, in such a case, sell his horses, let his house, turn off his servants, and make a thorough reform at once?" "it would be unnecessary for you to caution Mrs. Palmer and sir John against ever naming Mr. willoughby, or making the slightest allusion to what has passed, before my sister." "their attention to our comfort, their friendliness in every particular, is more than i can express." "she is not well, she has had a nervous complaint on her for several weeks." "could anything be so flattering as Mrs. ferrars's way of treating me yesterday? so exceedingly affable as she was!--you know how i dreaded the thoughts of seeing her;--but the very moment I was introduced, such an affability in her behaviour as really should seem to say, she had quite taken a fancy to me." her late conversation with her daughter-in-law had made her resolve on remaining at norland no longer than was unavoidable . to separate Edward and Elinor was as far from being her object as ever; and she wished to show Mrs. John dashwood how totally she disregarded her disapproval of the match . "my love i would ask them with all my heart, if it was in my power, but i had just settled within myself to ask the miss steeles to spend a few days with us." "i am sure you will like them; indeed, you DO like them, you know, very much already, and so does my mother; and they are such favourites with Harry!" "we all know how THAT will end:--they will wait a twelvemonth, and finding no good comes, will set down upon a curacy of fifty pounds a-year" "i must see what i can give them towards furnishing their house. two maids and two men, indeed!--as i talked of t'other day." "this little rectory CAN do no more than make mr. ferrars comfortable as a bachelor; it cannot enable him to marry." "what i am now doing indeed, seems nothing at all, since it can advance him so little towards what must be his principal, his only object of happiness." "Upon my soul it is,"--was his answer, with a warmth which brought all the former willoughby to her remembrance . "If that is all, you may be satisfied already,--for Marianne DOES--she has LONG forgiven you," he cried, in the same eager tone. "Then she has forgiven me before she ought to have done it." "your wife!--the letter was in your own hand-writing." "i will tell her all that is necessary to what may comparatively be called, your justification" "you have not explained to me the particular reason of your coming now" "No--to Combe Magna. i have business there; from thence to town in a day or two." "to judge from the colonel's spirits, however, you have not yet made him equally sanguine." "his age is only so much beyond hers as to make his character and principles fixed;--and his disposition, I am well convinced, is exactly the very one to make your sister happy" "my partiality does not blind me; certainly is not so handsome as willoughby--but at the same time, there is something much more pleasing in his countenance." "when do you write to Colonel Brandon, ma'am?" was an inquiry which sprung from her mind to have something going on . "i wrote to him, my love, last week, and rather expect to see, than to hear from him again." "do you call Colonel Brandon infirm?" asked Elinor . "you can hardly deceive yourself as to his having the use of his limbs!" "at this rate you must be in continual terror of MY decay," said her mother . "being very sure I have long lost your affections, i have thought myself at liberty to bestow my own on another, and have no doubt of being as happy with him as I once used to think I might be with you" "i scorn to accept a hand while the heart was another's, and it shall not be my fault if we are not always good friends, as our near relationship now makes proper" "your brother has gained my affections entirely, and as we could not live without one another, we are "I have burnt all your letters, and will return your picture the first opportunity" "please to destroy my scrawls--but the ring with my hair you are very welcome to keep" "worlds would not I have had a letter of hers seen by YOU in former days," said Edward . "how could i suppose, when she so earnestly, so warmly insisted on sharing my fate that any thing but the most disinterested affection was her inducement?" "i cannot comprehend on what motive she acted, or what fancied advantage it could be to her, to be fettered to a man for whom she had not the smallest regard"
    \\
    \bottomrule
    \end{tabular}
    \caption{
    Examples of decoded summaries of the full text of ``Sense and Sensibility", part 4.
    }
    \label{tab:decoded-summaries-books-part4}
\end{table*}

    \begin{table*}[th!]
    \centering
    \tiny
    \begin{tabular}{p{\linewidth}} 
    \toprule
    \textbf{Model Output: PEGASUS~\textsubscript{fine-tuned}} \\
    \midrule
    The Dashwood family has been settled in Sussex for many generations. Their estate is large, and their residence is at Norland Park, in the center of their property. The late owner of this estate was a single man, who lived to a very advanced age, and who for many years of his life had a constant companion and housekeeper in his sister. The Dashwoods are now settled at Barton with tolerable comfort to themselves. The house and the garden, with all the objects surrounding them, are now become familiar. Miss Dashwood has a delicate complexion, regular features, and a remarkably pretty figure. Her form, though not so correct as her sister's, in having the advantage of height, was more striking; and her face was so lovely, that when in the common cant of praise, she was called a beautiful girl, truth was less violently outraged than usually happens. Colonel Brandon's "partiality for Marianne, which had so early been discovered by his friends, now first became perceptible to Elinor, when it ceased to be noticed by them." She tells him that she has three reasons for disliking Colonel Brandon: he threatened her with rain when she wanted it to be fine, he has found fault with the hanging of her curricle, and she cannot persuade him to buy her brown mare. He says that he once knew a lady who in temper and mind greatly resembled your sister, who thought and judged like her, but who from an inforced change--from a series of unfortunate circumstances. When Sir John returns, he joins most heartily in the general regret on so unfortunate an event; concluding however by observing, that as they all got together, they must do something by way of being happy. He says that if they were one day to be your own, you would not be justified in what you have done. One evening in particular, about a week after Colonel Brandon left the country, his heart seemed more than usually open to every feeling of attachment to the objects around him. She flatters herself that even under the disadvantage of better rooms and a broader staircase, you will hereafter find your own house as faultless as you now do this. Mrs. Dashwood assures him that no alteration of the kind should be attempted. On their return from the park they find Willoughby's curricle and servant in waiting at the cottage, and Mrs. Dashwood is convinced that her conjecture had been just. She plays over every favorite song that she had been used to play to Willoughby, every air in which their voices had been oftenest joined, and sits at the instrument gazing on every line of music that he had written out for her. They are obliged to put an end to such an expectation. Charlotte laughs heartily to think that her husband could not get rid of her; and exultingly said, she did not care how cross he was to her, as they must live together. Mrs. Palmer's information regarding Willoughby is not very material, but any testimony in his favor, however small, is pleasing to her. The Palmers return to Cleveland the next day, and the two families at Barton are again left to entertain each other. The Miss Steeles, as she expected, had now all the benefit of these jokes, and in the eldest of them they raised a curiosity to know the name of the gentleman alluded to. Anne is the only person that knows of it, and she does me a great deal more harm than good, for I am in constant fear of her betraying me. Edward's visit to Plymouth, his melancholy state of mind, his dissatisfaction at his own prospects, his uncertain behaviour towards himself, the intimate knowledge of the Miss Steeles as to Norland and their family connections, which had often surprised her, the picture, the letter, the ring, formed altogether such a body of evidence, as overcame every fear of condemning him unfairly, and established as a fact, which no partiality could set aside, his ill-treatment of herself. Lady Middleton proposes a rubber of Casino to the others. No one makes any objection but Marianne, who with her usual inattention to the forms of general civility, exclaimed, "Your Ladyship will have the goodness to excuse ME--you know I detest cards." He has only two thousand pounds of his own; it would be madness to marry upon that, though for my own part, I could give up every prospect of more without a sigh. Mrs. Ferrars is a very proud woman, and in her first fit of anger upon hearing it, would very likely secure every thing to Robert, and the idea of that, for Edward's sake, frightens away all my inclination for hasty measures." He is not an ill-disposed young man, unless to be rather cold hearted and rather selfish is to be ill-posed: but he was, in general, well respected; for he conducted himself with propriety in the discharge of his ordinary duties. Had he married a more amiable woman, he might have been made still more respectable than he is, for he was very young when he married and very fond of his wife. Mrs. Palmer and two elderly ladies of Mrs. Jennings's acquaintance, whom she had met and invited in the morning, dine with them. He does not draw himself, but he has great pleasure in seeing the performances of other people, and he is by no means deficient in natural taste, though he has not had opportunities of improving it. They arrive at the place of destination, and as soon as the string of carriages before them allows, alighted, ascended the stairs, heard their names announced from one landing-place to another in an audible voice, and entered a room splendidly lit up, quite full of company, and insufferably hot. Mrs. Palmer's sympathy is shewn in procuring all the particulars in her power of the approaching marriage, and communicating them to Elinor. The calm and polite unconcern of Lady Middleton on the occasion was a happy relief to Elinor's spirits, oppressed as they often were by the clamorous kindness of the others. He asks her to introduce him to Sir John and Lady Middleton. Mrs. Ferrars is a little, thin woman, upright, and serious, even to sourness, in her aspect. Her complexion was sallow; and her features small, without beauty, and naturally without expression; but a lucky contraction of the brow had rescued her countenance from the disgrace of insipidity, by giving it the strong characters of pride and ill nature. She was not a woman of many words; for, unlike people in general, she proportioned them to the number of her ideas. The dinner is grand, the servants are numerous, and everything bespoke the Mistress's inclination for show, and the Master's ability to support it. They all look foolish; and Edward seems to have as great an inclination to walk out of the room again, as to advance farther into it. Lady Middleton is ashamed of doing nothing before them, and the flattery which Lucy was proud to think of and administer at other times, she feared they would despise her for offering. She says that Edward will marry Lucy; he will marry a woman superior in person and understanding to half her sex; and time and habit will teach him to forget that he ever thought another superior to HER. Mrs. Jennings is very warm in her praise of Edward's conduct, but only Elinor and Marianne understood its true merit. They only knew how little he had to tempt him to be disobedient, and how small was the consolation, beyond the consciousness of doing right, that could remain to him in the loss of friends and fortune. As soon as they return to the carriage, Mrs. Jennings is eager for information; but as Elinor wishes to spread as little as possible intelligence that had in the first place been so unfairly obtained, she confined herself to the brief repetition of such simple particulars, as she felt assured that Lucy, for the sake of her own consequence, would choose to have known. Mrs. Jennings is delighted with her gratitude, and only wondered that after hearing such a sentence, the Colonel should be able to take leave of them, as he immediately did, with the utmost sang-froid, and go away without making her any reply! Sir John's satisfaction in society is much more real. He is a blessing to all the juvenile part of the neighborhood for in summer he is for ever forming parties to eat cold ham and chicken out of doors, and in winter his private balls are numerous enough for any young lady who is not suffering under the unsatiable appetite of fifteen. The arrival of a new family in the country is always a matter of joy to him, and in every point of view he is charmed with the inhabitants he has now procured for his cottage at Barton. Mrs. Jennings, Lady Middleton's mother, is a "good-humoured, merry, fat, elderly woman, who talked a great deal, seemed very happy, and vulgar." She feels that his influence over her mind is heightened by circumstances which ought not in reason to have weight; by that person of uncommon attraction, that open, affectionate, and lively manner which it was no merit to possess; and by that still ardent love for Marianne. Colonel Brandon, the friend of Sir John, seemed no more adapted by resemblance of manner to be his friend than Lady Middleton was to be her wife, or Mrs. Jennings to be Lady Middleton's mother. His appearance however was not unpleasing, in spite of his being in the opinion of Marianne and Margaret an absolute old bachelor, for he was on the wrong side of five and thirty; but his face was not handsome, his countenance was sensible, and his address was particularly gentlemanlike. He is released from an entanglement which had long formed his misery, from a woman whom he had long ceased to love; and elevated at once to that security with another, which he must have thought of almost with despair. Edward is convinced that nothing could have been more natural than Lucy's conduct, nor more self-evident than the motive of it. The first month after their marriage is spent with their friend at the Mansion-house; from whence they could superintend the progress of the Parsonage. She was born to discover the falsehood of her own opinions, and to counteract, by her conduct, her most favourite maxims. She says that a woman of seven and twenty can never hope to feel or inspire affection again, and if her home be uncomfortable, or her fortune small, she can suppose that she might bring herself to submit to the offices of a nurse for the sake of the provision and security of a wife.
    \\
    \midrule
    \textbf{Model Output: PEGASUS~\textsubscript{zero-shot}} \\
    \midrule
    Had he been in his right senses, he could not have thought of such a thing as begging you to give away half your fortune from your own child." "If they were one day to be your own, Marianne, you would not be justified in what you have done." Dashwood again assured him that no alteration of the kind should be attempted. "Our dear Willoughby is now some miles from Barton, Elinor," said she, as she sat down to work, "and with how heavy a heart does he travel?" They were interrupted by the entrance of Margaret; and Elinor was then at liberty to think over the representations of her mother, to acknowledge the probability of many, and hope for the justice of all. "I am going into the village to see my horses," said he, "as you are not yet ready for breakfast; I shall be back again presently." *** I detest jargon of every kind, and sometimes I have kept my feelings to myself, because I could find no language to describe them in but what was worn and hackneyed out of all sense and meaning." And yet some of the plate would have been a very pleasant addition to our own stock here." Elinor was obliged to turn from her, in the middle of her story, to receive the rest of the party; Lady Middleton introduced the two strangers; Mrs. Palmer, on the contrary, who was strongly endowed by nature with a turn for being uniformly civil and happy, was hardly seated before her admiration of the parlour and every thing in it burst forth. "The rent of this cottage is said to be low; but we have it on very hard terms, if we are to dine at the park whenever any one is staying either with them, or with us." "Though we have seen him once or twice at my uncle's, it is rather too much to pretend to know him very well." It was broken by Lucy, who renewed the subject again by saying, with some hesitation, As she said this, she looked earnestly at Lucy, hoping to discover something in her countenance; perhaps the falsehood of the greatest part of what she had been saying; but Lucy's countenance suffered no change. "Oh," cried Miss Steele, looking significantly round at them, "I dare say Lucy's beau is quite as modest and pretty behaved as Miss Dashwood's." We seem so beset with difficulties on every side, that though it would make us miserable for a time, we should be happier perhaps in the end. "It will not be in my power to accept their invitation if they do." I thought so at the time; but you WOULD give him Cowper." The Middletons were to follow in about a week. Palmer appeared quite well, and I am commissioned to tell you, that you will certainly see her to-morrow." "How very odd!" said she, in a low and disappointed voice, as she turned away to the window. And Marianne was in spirits; happy in the mildness of the weather, and still happier in her expectation of a frost. Jennings's acquaintance to inform them of her being in town; and Marianne was all the time busy in observing the direction of the wind, watching the variations of the sky and imagining an alteration in the air. Elinor was alternately diverted and pained; but Marianne persevered, and saw every night in the brightness of the fire, and every morning in the appearance of the atmosphere, the certain symptoms of approaching frost. When you tell me to love him as a brother, I shall no more see imperfection in his face, than I now do in his heart." She then went away, walking on tiptoe out of the room, as if she supposed her young friend's affliction could be increased by noise. Jennings, with all her natural hilarity, burst forth again. Jennings, with a wine-glass, full of something, in her hand. With a letter in her outstretched hand, and countenance gaily smiling, from the persuasion of bringing comfort, she entered their room, saying, "Now, my dear, I bring you something that I am sure will do you good." Marianne was astonished to find how much the imagination of her mother and herself had outstripped the truth. "She is not well, she has had a nervous complaint on her for several weeks." Davies to be perfectly happy. Marianne had now been brought by degrees, so much into the habit of going out every day, that it was become a matter of indifference to her, whether she went or not: and she prepared quietly and mechanically for every evening's engagement, though without expecting the smallest amusement from any, and very often without knowing, till the last moment, where it was to take her. "They had already spent a week in this manner in Conduit Street, and Lady Middleton could not be displeased at their giving the same number of days to such near relations." Fanny paused a moment, and then, with fresh vigor, said, Dashwood and her daughters to begin their journey. Marianne got up and walked about the room. Jennings's curiosity and Elinor's too, that she would tell any thing WITHOUT being asked; for nothing would otherwise have been learnt. Elinor tried to talk of something else; but Miss Steele could not be kept beyond a couple of minutes, from what was uppermost in her mind. Jennings, who read it aloud with many comments of satisfaction and praise. Dashwood as soon as she could be assured that her visit would be no inconvenience; and as this message was answered by an invitation equally polite, her ladyship was introduced to them the next day. An opportunity was soon to be given to the Dashwoods of debating on the rest of the children, as Sir John would not leave the house without securing their promise of dining at the park the next day. Jennings, perhaps satisfied with the partial justification of her forebodings which had been found in their late alarm, allowed herself to trust in his judgment, and admitted, with unfeigned joy, and soon with unequivocal cheerfulness, the probability of an entire recovery. He was a blessing to all the juvenile part of the neighbourhood, for in summer he was for ever forming parties to eat cold ham and chicken out of doors, and in winter his private balls were numerous enough for any young lady who was not suffering under the unsatiable appetite of fifteen. But you have not explained to me the particular reason of your coming now, nor how you heard of her illness." On the contrary, with a mind and body alike strengthened by rest, she looked and spoke with more genuine spirit, anticipating the pleasure of Margaret's return, and talking of the dear family party which would then be restored, of their mutual pursuits and cheerful society, as the only happiness worth a wish. Edward was allowed to retain the privilege of first comer, and Colonel Brandon therefore walked every night to his old quarters at the Park; from whence he usually returned in the morning, early enough to interrupt the lovers' first tete-a-tete before breakfast. But if there should by any chance happen to be a woman who is single at seven and twenty, I should not think Colonel Brandon's being thirty-five any objection to his marrying HER."
    \\
    \bottomrule
    \end{tabular}
    \caption{
    Examples of decoded summaries of the full text of ``Sense and Sensibility", part 5.
    }
    \label{tab:decoded-summaries-books-part5}
\end{table*}

\end{document}